%% file: paper.tex
\documentclass[sigconf,nonacm]{aamas}
\input{resources/aamas-compat.tex}
\input{resources/preamble.tex}
\setcopyright{none}
\title[Do More Agents Help?]{Do More Agents Help? Evaluating Accuracy and Cost in LLM Multi-Agent Workflows}
\author{Yuhang Fu}
\authornote{Work was done during Yuhang's internship at Westlake University.}
\affiliation{\institution{Beijing University of Posts and Telecommunications}\country{China}}
\email{fuzi1fuzi1@bupt.edu.cn}
\author{Ruishan Fang}
\affiliation{\institution{Zhejiang University}\country{China}}
\affiliation{\institution{Westlake University}\country{China}}
\email{fangruishan@westlake.edu.cn}
\author{Jiaqi Shao}
\affiliation{\institution{Duke Kunshan University}\country{China}}
\affiliation{\institution{Hong Kong University of Science and Technology}\country{China}}
\email{js1139@duke.edu}
\author{Huiyu Zheng}
\affiliation{\institution{Zhejiang University of Technology}\country{China}}
\email{221124120277@zjut.edu.cn}
\author{Zhengtao Zhu}
\affiliation{\institution{Zhejiang University of Technology}\country{China}}
\email{zhentaozhu@zjut.edu.cn}
\author{Bing Luo}
\affiliation{\institution{Duke Kunshan University}\country{China}}
\email{bing.luo@dukekunshan.edu.cn}
\author{Zhenyu Wen}
\affiliation{\institution{Zhejiang University of Technology}\country{China}}
\email{wenluke427@gmail.com}
\author{Tao Lin}
\affiliation{\institution{Westlake University}\country{China}}
\email{lintao@westlake.edu.cn}
\renewcommand{\shortauthors}{Fu et al.}
\input{resources/abstract.tex}

\definecolor{buptcolor}{HTML}{68139A}
\definecolor{westlakecolor}{HTML}{009E73}
\definecolor{zjucolor}{HTML}{2878B5}
\definecolor{dukecolor}{HTML}{D86913}
\definecolor{hkustcolor}{HTML}{009AC8}
\definecolor{zjutcolor}{HTML}{B63250}
\newcommand{\buptmark}{\textcolor{buptcolor}{\ensuremath{\spadesuit}}}
\newcommand{\westlakemark}{\textcolor{westlakecolor}{\ensuremath{\blacktriangle}}}
\newcommand{\zjumark}{\textcolor{zjucolor}{\ensuremath{\clubsuit}}}
\newcommand{\dukemark}{\textcolor{dukecolor}{\ensuremath{\bigstar}}}
\newcommand{\hkustmark}{\textcolor{hkustcolor}{\ensuremath{\blacklozenge}}}
\newcommand{\zjutmark}{\textcolor{zjutcolor}{\ensuremath{\blacksquare}}}
\newcommand{\authormarks}[1]{\textsuperscript{#1}}

\newcommand{\arxivauthorblock}{%
  \begin{minipage}{\textwidth}
    \fontfamily{phv}\selectfont
    \setlength{\tabcolsep}{0pt}
    \begin{tabular}{@{}p{0.25\linewidth}p{0.25\linewidth}p{0.25\linewidth}p{0.25\linewidth}@{}}
      \textbf{Yuhang Fu}\authormarks{\buptmark\,$\ddagger$}\textsuperscript{*} &
      \textbf{Ruishan Fang}\authormarks{\zjumark\,\westlakemark\,$\ddagger$} &
      \textbf{Jiaqi Shao}\authormarks{\dukemark\,\hkustmark} &
      \textbf{Huiyu Zheng}\authormarks{\zjutmark}\\[0.55em]
      \textbf{Zhengtao Zhu}\authormarks{\zjutmark} &
      \textbf{Bing Luo}\authormarks{\dukemark\,$\dagger$} &
      \textbf{Zhenyu Wen}\authormarks{\zjutmark\,$\dagger$} &
      \textbf{Tao Lin}\authormarks{\westlakemark\,$\dagger$}
    \end{tabular}
    \par\vspace{0.9em}
    {\fontsize{9.5}{12}\selectfont
    \begin{tabular}{@{}p{0.64\linewidth}p{0.36\linewidth}@{}}
      \textcolor{buptcolor}{\buptmark\,Beijing University of Posts and Telecommunications} &
      \textcolor{westlakecolor}{\westlakemark\,Westlake University}\\[0.25em]
      \textcolor{zjucolor}{\zjumark\,Zhejiang University} &
      \textcolor{dukecolor}{\dukemark\,Duke Kunshan University}\\[0.25em]
      \textcolor{hkustcolor}{\hkustmark\,Hong Kong University of Science and Technology} &
      \textcolor{zjutcolor}{\zjutmark\,Zhejiang University of Technology}
    \end{tabular}}
    \par\vspace{0.8em}
    {\fontsize{9}{12}\selectfont
    \textsuperscript{$\ddagger$}Equal contribution.\quad
    \textsuperscript{$\dagger$}Corresponding authors:\\[0.25em]
    \href{mailto:bing.luo@dukekunshan.edu.cn}{bing.luo@dukekunshan.edu.cn}\quad
    \href{mailto:wenluke427@gmail.com}{wenluke427@gmail.com}\quad
    \href{mailto:lintao@westlake.edu.cn}{lintao@westlake.edu.cn}}
  \end{minipage}
}

\makeatletter
\def\@titlefont{\fontfamily{phv}\selectfont\bfseries\fontsize{20}{24}\selectfont}
\def\@mkauthors@iii{%
  \global\setbox\mktitle@bx=\vbox{%
    \unvbox\mktitle@bx\par\noindent
    {\fontsize{12}{15}\selectfont\arxivauthorblock}\par\bigskip}}
\makeatother
\hypersetup{
  pdftitle={Do More Agents Help? Evaluating Accuracy and Cost in LLM Multi-Agent Workflows},
  pdfauthor={Yuhang Fu, Ruishan Fang, Jiaqi Shao, Huiyu Zheng, Zhengtao Zhu, Bing Luo, Zhenyu Wen, Tao Lin},
  colorlinks=true,linkcolor=blue,citecolor=blue,urlcolor=blue
}
\begin{document}
\maketitle
\fancypagestyle{preprint}{\fancyhf{}\fancyfoot[C]{\thepage}\renewcommand{\headrulewidth}{0pt}}
\pagestyle{preprint}
\thispagestyle{preprint}
\input{resources/main.tex}
\bibliographystyle{ACM-Reference-Format}
\bibliography{resources/reference}
\label{arxiv:references-end}
\clearpage
\onecolumn
\newgeometry{margin=1in}
\fontsize{10}{12}\selectfont
\emergencystretch=1em
\appendix
\renewcommand{\thetable}{S\arabic{table}}
\renewcommand{\thefigure}{S\arabic{figure}}
\renewcommand{\theHtable}{appendix.\arabic{table}}
\renewcommand{\theHfigure}{appendix.\arabic{figure}}
\setcounter{table}{0}
\setcounter{figure}{0}
\section*{Supplementary Material}
\input{resources/appendix.tex}

\end{document}

%% file: resources/aamas-compat.tex
\makeatletter
\def\AAMAS@missingeditors{\fi}
\ifx\acmConference@editors\AAMAS@missingeditors
  \expandafter\@firstofone
\else
  \expandafter\@gobble
\fi
{\fi}
\makeatother

%% file: resources/preamble.tex
\usepackage{multirow,array,makecell,microtype,tikz,longtable}
\usetikzlibrary{arrows.meta}

\usepackage{balance}
\makeatletter
\AddToHook{cmd/bibitem/before}{%
  \ifnum\value{page}=\getpagerefnumber{arxiv:references-end}\relax
    \if@firstcolumn\balance\fi
  \fi}
\makeatother

%% file: resources/abstract.tex
\begin{abstract}
    Choosing whether to replace a single-agent controller requires evidence on both task performance and inference expenditure.
    We present BenchAgent, a shared evaluation framework, and compare six multi-agent workflow configurations against its single-agent reference, BenchAgent Core, across ten reasoning, coding, and tool-use benchmarks.
    All configurations use GPT-4.1 and common tool and scoring interfaces; Core executions share the same step limits. We measure accuracy and recorded token use over three runs on fixed task sets, with 3,302 instances per configuration per run.
    EvoAgent reaches 79.70\% mean pooled accuracy versus 76.65\% for Core and exceeds Core in all three runs; the other five MAS trail by 1.71--8.76 percentage points.
    LLM-Debate gains 9.67 points on HumanEval but loses 1.96 points in pooled accuracy while consuming 1.39 times the tokens of Core.
    ChatEval gains 22.25 points on IFEval but trails Core in pooled accuracy with 4.34 times its token consumption.
    The broad comparison and targeted analyses of retained responses provide evidence for workflow selection and identify answer preservation and resource allocation as directions for subsequent multi-agent optimization.
\end{abstract}

%% file: resources/main.tex
\section{Introduction}
\label{sec:introduction}
Multi-agent workflows offer several ways to organize the capabilities of a language model: debate candidate answers, exchange role-specific messages, delegate tool use, or evolve agent instructions~\citep{du2023multiagentdebate,li2023camel,wu2023autogen,shen2023hugginggpt,yuan2024evoagent}.
A workflow may improve accuracy on one benchmark yet perform worse across a broader task suite. Even a higher aggregate score can hide regressions on particular tasks.
Workflows with similar accuracy may also differ in token consumption.
We examine benchmark-level results alongside pooled accuracy and recorded token use to identify where replacing a single-agent controller brings gains, where it incurs costs, and which behaviors merit further optimization.

Recent studies examine how workflow structure affects performance and computational cost. Xu et al.~\citep{xu2026rethinking} compare homogeneous workflows with single-LLM executions that retain their workflow logic. Kim et al.~\citep{kim2025scaling} compare canonical coordination architectures across interactive tasks and model families.
We complement these studies by comparing workflow designs normalized to a common execution substrate: Jarvis, LLM-Debate, EvoAgent, AutoGen, CAMEL, and ChatEval, each compared with a common controller, BenchAgent Core. Across ten fixed task sets, we report gains, regressions, and token expenditure, then use retained responses to identify concrete directions for improving these configurations.

BenchAgent provides common benchmark inputs, model settings, tool interfaces, answer contracts, scoring rules, and token accounting across workflows.
We evaluate BenchAgent Core, a single task-level controller, and six MAS configurations on ten benchmarks spanning mathematical and general reasoning, reading comprehension, coding, instruction following, and retrieval.
The primary study uses GPT-4.1 and three runs on fixed task sets, with 3,302 instances per configuration per run.
We also report single-run GAIA results and auxiliary backend checks, along with a separate case study of an external runtime.

\begin{figure*}[t]
    \centering
    \includegraphics[width=0.84\textwidth]{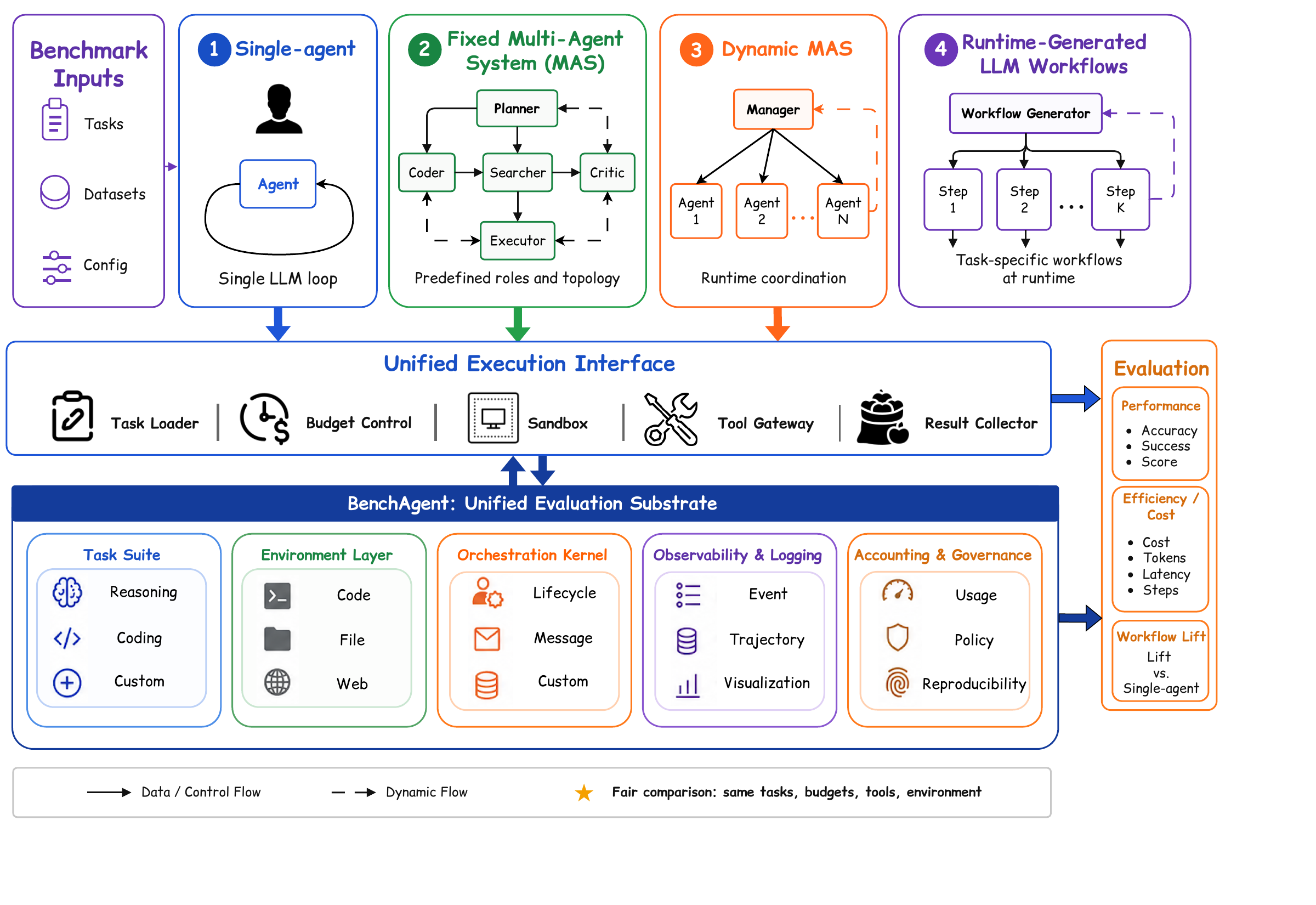}
    \caption{\textbf{BenchAgent workflow and evaluation overview.} The diagram illustrates general workflow organizations and the shared execution substrate. The primary comparison evaluates Core, fixed MAS, and EvoAgent's evolving prompts with shared Core step limits; the external CC-workflow is evaluated separately as a supplementary case study.}
    \Description{General single-agent, fixed multi-agent, and dynamic workflows above a shared execution substrate; a separate runtime-generated workflow illustrates the external case-study category.}
    \label{fig:paradigm_centered_setup}
\end{figure*}

Figure~\ref{fig:paradigm_centered_setup} summarizes the comparison protocol.
We report accuracy and token use for individual benchmarks and the pooled suite, using BenchAgent Core as the single-agent reference throughout.
The study makes three contributions:

\noindent\textbf{A shared framework for workflow comparison.} BenchAgent places normalized multi-agent implementations and a single-agent controller under common task, tool, answer, scoring, and accounting interfaces. This supports consistent measurement of complete configurations and provides a common basis for evaluating subsequent workflow changes.

\noindent\textbf{Broad evidence across seven workflows and ten benchmarks.} We compare six MAS configurations against BenchAgent Core over three runs, quantifying benchmark-specific gains, regressions, and pooled accuracy--token trade-offs. The resulting profiles show how each workflow's task strengths relate to its aggregate performance and cost, providing evidence for workflow selection and further optimization.

\noindent\textbf{Targeted evidence for workflow optimization.} Analyses of 371 parseable ChatEval GSM8K tasks and 163 matched Jarvis--ChatEval GAIA tasks relate outcomes to token expenditure, answer changes, and execution-limit signals. These observations identify answer preservation and resource allocation as concrete directions for subsequent optimization; they do not establish the causal effects of individual mechanisms.

\section{Related Work}
\subsection{Agent Workflow Design}
ReAct, Toolformer, WebGPT, and SWE-agent show that a single instrumented controller can already serve as a competitive workflow~\citep{yao2022react,schick2023toolformer,nakano2021webgpt,yang2024sweagent}.
Recent studies compare multi-agent workflows with single-LLM executions that preserve workflow logic~\citep{xu2026rethinking} and examine coordination architectures across tasks and model families~\citep{kim2025scaling}. Our comparison covers six named implementations and a common controller on ten benchmarks, combining accuracy and token measurements with targeted response analyses.

Multi-agent systems make the division of labor explicit.
HuggingGPT/JARVIS organizes a controller and task executors through four stages: task planning, model selection, task execution, and response generation~\citep{shen2023hugginggpt}.
CAMEL and MetaGPT define agent roles and communication procedures~\citep{li2023camel,hong2023metagpt}, AutoGen provides a conversable-agent substrate for fixed or adaptive interaction patterns~\citep{wu2023autogen}, and debate-style systems aggregate multiple proposals before finalization~\citep{du2023multiagentdebate,chan2023chateval}.
Li~\emph{et al.}~\citep{li2024moreagents} show that sampling-and-voting improves several LLM benchmarks; DSPy~\citep{khattab2023dspy} instead optimizes LM pipelines through compilation.
Dynamic and evolving systems search over or mutate the workflow structure within a designer-specified space, as in AgentVerse, EvoAgent, ADAS, AFlow, and Magentic-One~\citep{chen2023agentverse,yuan2024evoagent,hu2024adas,zhang2024aflow,fourney2024magenticone}; MaAS extends this by searching over query-dependent agent topologies via an agentic supernet~\citep{zhang2025maas}; MASPO further co-optimizes agent prompts jointly across the pipeline~\citep{wang2026maspo}.
Cemri~\emph{et al.}~\citep{cemri2025maswhyfail} catalog MAS failure modes attributable to coordination design; Kim~\emph{et al.}~\citep{kim2025scaling} further argue that net workflow gain requires coordination benefit to exceed communication overhead.
We compare complete fixed and evolving configurations through shared execution and measurement interfaces, and examine the external-runtime workflow separately.

\subsection{Agent Evaluation Frameworks}
AgentBench, GAIA, AgentBoard, WebArena, SWE-bench, ToolBench, and Silo-Bench~\citep{liu2023agentbench,mialon2023gaia,ma2024agentboard,zhou2023webarena,jimenez2024swebench,qin2023toolbench,zhang2026silobench} extend agent evaluation to interaction, tool use, web/software tasks, analytical traces, and distributed coordination.
Harbor~\citep{harbor_framework} evaluates and optimizes sandboxed agents and models in containerized environments.
BenchAgent uses common execution and tool interfaces, usage accounting, and trajectory logging to compare workflows on existing benchmarks.
Appendix Table~\ref{tab:appendix_shared_interfaces} specifies the shared interfaces and workflow-dependent components.

\subsection{Runtime-Generated Workflows}
Recent engineering systems push part of workflow design into the execution loop.
LangGraph~\citep{langgraph_docs} and CrewAI~\citep{crewai_docs} provide graph orchestration and role-based crews; Anthropic~\citep{anthropic_building_effective_agents,anthropic_multiagent_research_system,anthropic_context_engineering} documents subagent creation, context-window isolation, and permission-scoped delegation patterns instantiated in Claude Code~\citep{anthropic_claude_code_product,anthropic_claude_code_overview,anthropic_claude_code_subagents}; similar agentic coding interfaces appear in OpenAI Codex and OpenCode~\citep{openai_codex_product,opencode_product}.
These systems are documented primarily through official documentation rather than peer-reviewed publications; we cite them as engineering descriptions, not empirical evidence.
Our supplementary GAIA case study examines the CC-workflow with a GPT-4.1 backend, reporting its accuracy, token use, and retained execution traces alongside BenchAgent reference results.

\section{Evaluation Protocol}
\label{sec:evaluation_protocol}
\subsection{Workflow Lift as the Measurement Target}
We evaluate agent systems as \emph{workflow organizations}, not isolated products.
For a benchmark instance \(x\), a workflow \(w\) produces a final answer \(\hat{y}=w(x)\) and, when available, an execution trace \(\tau_w(x)\) containing model calls, tool calls, messages, artifacts, and termination events.
A task evaluator \(E(x,\hat{y}) \in \{0,1\}\) measures success, while a cost summary \(c(\tau_w)\) records usage signals such as tokens, latency, tool calls, and delegation structure.
The comparison therefore asks not only whether \(\hat{y}\) is correct, but also what trace \(\tau_w\) produced it.

We distinguish workflow categories by their task-level controllers, role prompts, communication structure, and tool scopes. Shared tool services, including the search helper, are distinguished from workflow-level roles:
\begin{itemize}
    \item \textbf{Single-agent workflow.} Uses one persistent task-level controller over the shared tool interface, which can delegate web research to a search helper.
    \item \textbf{Fixed MAS.} Uses predefined role types and communication rules, such as solver--critic--aggregator or debate-style handoffs. The stage-level procedure is fixed, while task executors can be instantiated for the current task, as in the Jarvis workflow.
    \item \textbf{Evolving MAS.} Searches over a predefined configuration family. The tested EvoAgent evolves agent prompts while retaining the shared execution substrate.
    \item \textbf{Runtime-generated workflow.} Instantiates task-specific agents and assigns roles, context, tool scopes, or routing rules during execution.
\end{itemize}
Table~\ref{tab:appendix_workflow_taxonomy} gives the full terminology map.

\begin{definition}[Workflow lift]
    \label{def:workflow_lift}
    We measure \emph{workflow lift} as the change in accuracy and cost relative to the BenchAgent Core single-agent configuration when it is replaced by a fixed or evolving MAS workflow, with the base model, benchmark loader, Core step limits, tool interface, answer contract, evaluator, and accounting held constant.
\end{definition}

This comparison measures complete workflow configurations under shared execution interfaces. Section~\ref{subsec:reporting_protocol} describes how we aggregate results across runs.

\subsection{Shared Evaluation Setup}
\label{subsec:comparison_setup}
BenchAgent standardizes benchmark loading, input and answer formatting, runtime control, tool access, usage and trajectory logging, and evaluator calls.
The six MAS instantiate published workflow designs on this common substrate. Normalization controls differences in model backends, tool environments, and evaluation procedures across the original systems; role prompts, communication procedures, and role-level tool allocation define the workflow configurations being compared.
The measurement target is these configurations within BenchAgent, rather than reproduction of each original system in its native environment.
Table~\ref{tab:main_shared_interface_map} summarizes these shared and variable components; Appendix Table~\ref{tab:appendix_shared_interfaces} gives the full map.

Each Core execution allows up to 15 manager steps and 10 search-helper steps. Each workflow determines the number of Core invocations and how role messages and candidate answers are combined. Recorded token expenditure is an outcome of this execution under shared step limits, rather than a workflow-level token quota.

The supplementary CC case study uses an external runtime; Section~\ref{subsec:cc_case_study} describes its settings and interpretation separately.

\begin{table}[t]
    \centering
    \caption{Shared interfaces and workflow-specific behavior in the MAS accuracy and cost comparison.}
    \label{tab:main_shared_interface_map}
    \footnotesize
    \setlength{\tabcolsep}{3pt}
    \begin{tabular}{@{}>{\raggedright\arraybackslash}p{0.24\linewidth}>{\raggedright\arraybackslash}p{0.32\linewidth}>{\raggedright\arraybackslash}p{0.38\linewidth}@{}}
        \toprule
        Shared interface & Wrapper variation & Comparison enabled \\
        \midrule
        Input and evaluator & Role prompts and answer aggregation & Accuracy on the same tasks and scoring rules \\
        Tool registry & Role-level tool choices & Tool use under matched availability \\
        Usage accounting & Number and sequence of model calls & Token expenditure alongside accuracy \\
        Trajectory logging & Messages and handoffs & Process differences behind final scores \\
        \bottomrule
    \end{tabular}
\end{table}

\paragraph{Extending BenchAgent.}
Adding a benchmark typically requires two benchmark-specific components: a data adapter that maps source records to BenchAgent's normalized instance schema and an evaluator that extracts, normalizes, and scores the final answer.
Once registered, the benchmark reuses the common runner, tool interface, trajectory logger, and accounting pipeline, so existing workflows do not require benchmark-specific changes to their execution logic.

\begin{figure*}[t]
    \centering
    \includegraphics[width=0.80\textwidth]{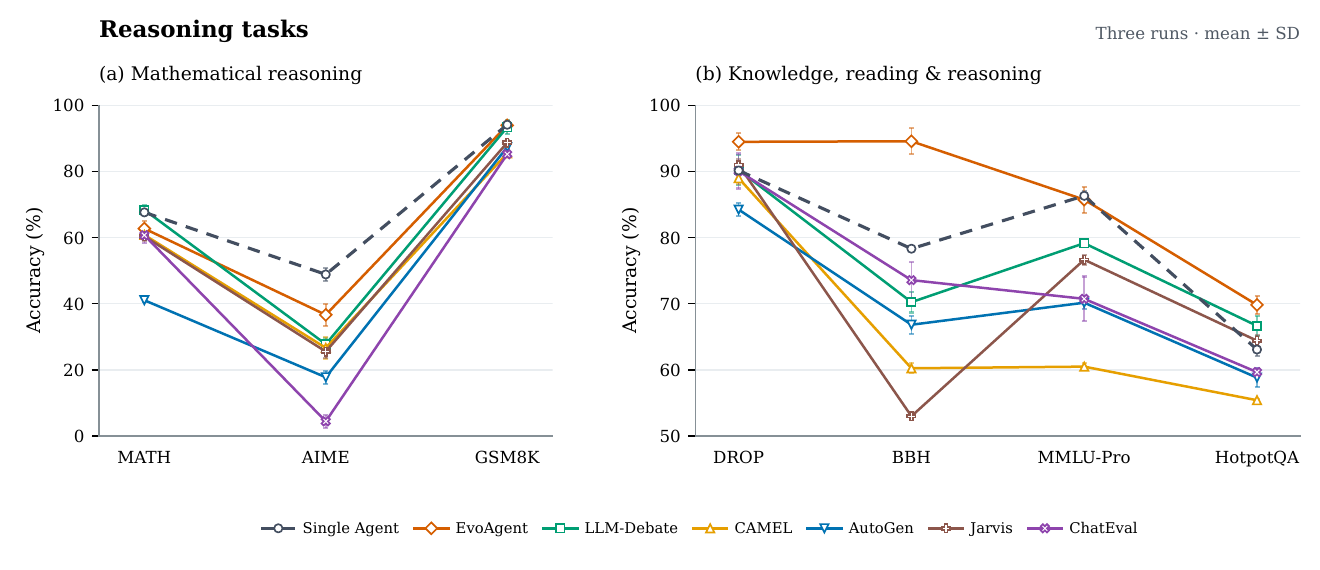}
    \caption{
        \textbf{Reasoning-task accuracy across seven workflow configurations.} Points show mean accuracy over three runs; error bars show $\pm$ one sample standard deviation. BenchAgent Core is the gray dashed single-agent line. Panel (a) covers mathematical reasoning on a 0--100\% axis; panel (b) covers knowledge, reading, and general reasoning on a 50--100\% axis. Lines connect categorical benchmark results as visual guides. All points and error bars are shown in full.
    }
    \Description{Two panels compare three-run accuracy for seven workflow configurations across seven reasoning benchmarks, with standard-deviation error bars.}
    \label{fig:broad_reasoning_categories}
\end{figure*}

\begin{table*}[!t]
    \centering
    \caption{
        \textbf{Accuracy and token use of six MAS configurations relative to BenchAgent Core.} Single Agent denotes Core. Accuracy reports the mean $\pm$ sample standard deviation of three runs (percent and percentage points, respectively). Pooled Acc. sums correct answers over 3,302 instances within each run before aggregation. Pooled Tok. (avg) reports the three-run average token usage per instance, weighted by benchmark size. Section~\ref{subsec:reporting_protocol} gives the calculation.
    }
    \label{tab:main_exp1_overall}
    \setlength{\tabcolsep}{3.8pt}
    \renewcommand{\arraystretch}{1.08}
    \small
    \resizebox{\textwidth}{!}{
        \begin{tabular}{lccccccc}
            \toprule
            \textbf{Benchmark / Metric} & \textbf{Single Agent}          & \textbf{EvoAgent}              & \textbf{LLM-Debate}            & \textbf{CAMEL}                 & \textbf{AutoGen}               & \textbf{Jarvis}                & \textbf{ChatEval}             \\
            \midrule
            MATH & 67.67{\scriptsize $\pm$1.01}\% & 62.75{\scriptsize $\pm$2.41}\% & 68.42{\scriptsize $\pm$1.44}\% & 60.83{\scriptsize $\pm$1.51}\% & 41.08{\scriptsize $\pm$0.52}\% & 60.42{\scriptsize $\pm$1.26}\% & 60.75{\scriptsize $\pm$2.41}\% \\
            AIME & 48.89{\scriptsize $\pm$1.92}\% & 36.67{\scriptsize $\pm$3.33}\% & 27.78{\scriptsize $\pm$1.92}\% & 26.67{\scriptsize $\pm$3.33}\% & 17.78{\scriptsize $\pm$1.92}\% & 25.56{\scriptsize $\pm$1.92}\% & 4.44{\scriptsize $\pm$1.92}\% \\
            GSM8K & 94.17{\scriptsize $\pm$1.04}\% & 94.00{\scriptsize $\pm$1.09}\% & 93.42{\scriptsize $\pm$2.13}\% & 85.50{\scriptsize $\pm$1.09}\% & 87.42{\scriptsize $\pm$0.52}\% & 88.75{\scriptsize $\pm$0.90}\% & 85.25{\scriptsize $\pm$0.25}\% \\
            DROP & 90.17{\scriptsize $\pm$1.28}\% & 94.50{\scriptsize $\pm$1.30}\% & 90.25{\scriptsize $\pm$2.29}\% & 89.00{\scriptsize $\pm$0.87}\% & 84.25{\scriptsize $\pm$1.00}\% & 91.00{\scriptsize $\pm$0.87}\% & 90.08{\scriptsize $\pm$2.65}\% \\
            BBH & 78.33{\scriptsize $\pm$0.14}\% & 94.58{\scriptsize $\pm$1.94}\% & 70.25{\scriptsize $\pm$1.56}\% & 60.25{\scriptsize $\pm$0.75}\% & 66.83{\scriptsize $\pm$1.38}\% & 53.00{\scriptsize $\pm$0.50}\% & 73.58{\scriptsize $\pm$2.67}\% \\
            MMLU-Pro & 86.33{\scriptsize $\pm$0.76}\% & 85.75{\scriptsize $\pm$1.95}\% & 79.17{\scriptsize $\pm$0.38}\% & 60.50{\scriptsize $\pm$0.50}\% & 70.17{\scriptsize $\pm$0.95}\% & 76.67{\scriptsize $\pm$0.72}\% & 70.75{\scriptsize $\pm$3.38}\% \\
            HumanEval & 84.99{\scriptsize $\pm$0.44}\% & 81.17{\scriptsize $\pm$0.44}\% & 94.66{\scriptsize $\pm$0.76}\% & 88.04{\scriptsize $\pm$1.17}\% & 61.07{\scriptsize $\pm$1.32}\% & 88.80{\scriptsize $\pm$0.44}\% & 85.50{\scriptsize $\pm$0.76}\% \\
            MBPP & 68.52{\scriptsize $\pm$0.17}\% & 72.24{\scriptsize $\pm$0.90}\% & 72.92{\scriptsize $\pm$2.06}\% & 74.49{\scriptsize $\pm$0.59}\% & 73.80{\scriptsize $\pm$0.90}\% & 75.07{\scriptsize $\pm$1.06}\% & 76.15{\scriptsize $\pm$1.11}\% \\
            HotpotQA & 63.08{\scriptsize $\pm$1.04}\% & 69.83{\scriptsize $\pm$1.38}\% & 66.67{\scriptsize $\pm$1.53}\% & 55.42{\scriptsize $\pm$0.14}\% & 58.83{\scriptsize $\pm$1.38}\% & 64.42{\scriptsize $\pm$0.80}\% & 59.67{\scriptsize $\pm$0.58}\% \\
            IFEval & 63.08{\scriptsize $\pm$0.29}\% & 65.58{\scriptsize $\pm$1.26}\% & 53.17{\scriptsize $\pm$2.18}\% & 60.42{\scriptsize $\pm$1.38}\% & 67.58{\scriptsize $\pm$0.72}\% & 43.00{\scriptsize $\pm$0.25}\% & 85.33{\scriptsize $\pm$0.95}\% \\
            \midrule
            Pooled Acc. & 76.65{\scriptsize $\pm$0.32}\% & 79.70{\scriptsize $\pm$0.26}\% & 74.69{\scriptsize $\pm$0.41}\% & 68.59{\scriptsize $\pm$0.16}\% & 67.89{\scriptsize $\pm$0.54}\% & 69.32{\scriptsize $\pm$0.29}\% & 74.94{\scriptsize $\pm$0.96}\% \\
            Pooled Tok. (avg) & 23712.97 & 30036.80 & 32987.40 & 7724.76 & 11634.54 & 8654.41 & 103022.69 \\

            \bottomrule
        \end{tabular}
    }
    \renewcommand{\arraystretch}{1.0}
\end{table*}

\begin{figure*}[t]
    \centering
    \includegraphics[width=0.88\textwidth]{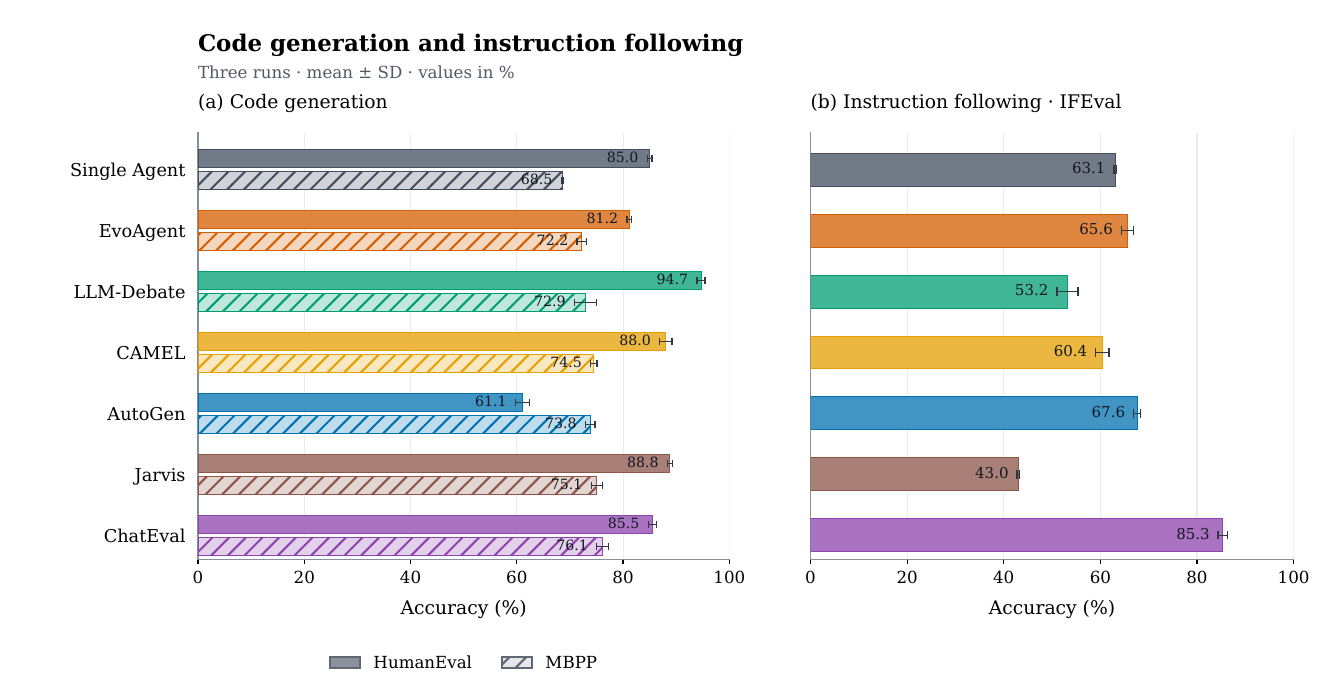}
    \caption{
        \textbf{Code-generation and instruction-following accuracy.} Bars show three-run mean accuracy and error bars show $\pm$ one sample standard deviation. Panel (a) pairs HumanEval (upper solid bar) and MBPP (lower hatched bar) for each method; panel (b) reports IFEval. Both accuracy axes span 0--100\%, and method colors and row order are shared. Labels show mean accuracy to one decimal place; Table~\ref{tab:main_exp1_overall} reports the full mean and standard deviation to two decimal places.
    }
    \Description{Horizontal bars compare seven workflow configurations on HumanEval, MBPP, and IFEval, showing three-run mean accuracy and standard deviations.}
    \label{fig:broad_code_instruction_categories}
\end{figure*}

\subsection{Metrics and Aggregation}
\label{subsec:reporting_protocol}
We report final-answer accuracy and recorded LLM token usage. Accuracy is the fraction of evaluated instances that satisfy the benchmark-specific success criteria in Appendix Table~\ref{tab:appendix_scoring_rules}. Pooled Acc. divides the total correct count across the ten benchmarks by their combined 3,302 instances, so larger splits receive more weight. GAIA overall pools manually reviewed correctness labels over 165 tasks.

We evaluate each broad-suite configuration three times on the same task sets and report mean accuracy and sample standard deviation across runs. We compute pooled accuracy within each run before averaging. GAIA and auxiliary backend results are single-run measurements. Token usage includes recorded input and output tokens across workflow calls; Pooled Tok. averages the per-instance benchmark means across three runs, weighted by split size. Appendix~\ref{subsec:appendix_scoring} gives the detailed aggregation rules.

Appendix Tables~\ref{tab:appendix_broad_correct_counts}--\ref{tab:appendix_broad_tokens} report correct counts for each run and mean token usage for each benchmark. Selected execution traces show how workflows use tools, exchange messages, and delegate tasks.

\section{Experiments}
\label{sec:experiments}
The primary evaluation compares BenchAgent workflow configurations across ten benchmarks using three runs under the protocol of Section~\ref{sec:evaluation_protocol}.
We also report single-run GAIA results and auxiliary backend checks for BenchAgent-compatible workflows.
The final subsection adds an external Claude Code case study on GAIA.

\subsection{Experimental Setup}
\label{subsec:exp_setup}
\noindent\textbf{Datasets.}
The broad-suite comparison uses frozen, processed subsets from MASArena for ten benchmarks: MATH, AIME, GSM8K, DROP, BBH, MMLU-Pro, HumanEval, MBPP, HotpotQA, and IFEval. We sample 400 instances with seed 42 from processed files containing more than 400 instances and otherwise evaluate the full file. Appendix~\ref{subsec:appendix_protocol_fidelity} lists dataset sources and sampling details.
GAIA uses all 165 tasks in the 2023 validation split~\citep{mialon2023gaia}, covering retrieval, file inspection, and tool use. The three levels contain 53, 86, and 26 tasks; overall accuracy weights levels by these sizes.

\noindent\textbf{Tool allocation.}
The available tool registry is fixed for each benchmark across BenchAgent workflows; role-level tool allocation and tool selection depend on the workflow.
HotpotQA uses the full BenchAgent tool registry. The other nine broad-suite benchmarks allow only Python execution in addition to final-answer submission.
On GAIA, BenchAgent workflows use the full registry. The CC case study provides the corresponding classes of tool capabilities. Appendix Table~\ref{tab:appendix_tool_regimes} lists the tool regimes.

\noindent\textbf{Compared systems.}
BenchAgent Core is the single-agent reference, sharing the loader, evaluator, and tool registry with all MAS wrappers.
Fixed MAS baselines include Jarvis~\citep{shen2023hugginggpt}, LLM-Debate~\citep{du2023multiagentdebate}, AutoGen~\citep{wu2023autogen}, CAMEL~\citep{li2023camel}, and ChatEval~\citep{chan2023chateval}; EvoAgent~\citep{yuan2024evoagent} represents the evolving MAS category.
Implementation notes appear in Appendix Tables~\ref{tab:appendix_reproduced_mas_contracts}--\ref{tab:appendix_evoagent_contract}; Table~\ref{tab:appendix_baseline_fidelity} reports auxiliary transfer sanity checks.

\noindent\textbf{Model and infrastructure.}
For the broad suite and the main GAIA comparison, we use GPT-4.1 (\texttt{gpt-4.1-2025-04-14})~\citep{openai_gpt41_model}.
We use Qwen3-32B for an auxiliary GAIA workflow comparison and GLM-5 for a separate calibration of Core (Appendix Table~\ref{tab:appendix_glm5_gaia_generalization}). The appendix gives further details on dataset sampling, tools, and execution traces.

\subsection{Broad Benchmark Results}
\label{subsec:broad_bench_results}
Table~\ref{tab:main_exp1_overall} reports three-run accuracy and mean token use under the shared model, tool, and evaluation protocol. We compare each workflow with BenchAgent Core both on individual benchmarks and across the pooled suite.

\noindent\textbf{The pooled result repeats across all three runs.} Mean pooled accuracy is 76.65\% for Core ($\pm$0.32 points) and 79.70\% for EvoAgent ($\pm$0.26 points). EvoAgent gains +3.03, +3.15, and +2.97 points over Core in the three runs. The other five MAS are below Core in every run, with mean pooled accuracies from 67.89\% to 74.94\%.

\noindent\textbf{Higher token consumption does not guarantee higher accuracy.} LLM-Debate and ChatEval average 32,987.40 and 103,022.69 tokens per instance, respectively, versus 23,712.97 for Core. They use 1.39 and 4.34 times as many tokens but have lower mean pooled accuracy. EvoAgent gains +3.05 points in mean pooled accuracy while using 1.27 times Core's recorded tokens. CAMEL and Jarvis use fewer tokens and have lower pooled accuracy. These token differences include the calls and accumulated histories induced by each workflow under shared Core step limits.

\noindent\textbf{Local gains need not yield a pooled advantage.} EvoAgent and LLM-Debate each exceed Core's mean accuracy on five of ten benchmarks, but their pooled differences are +3.05 and $-1.96$ points. The magnitude and task distribution of gains therefore matter alongside the number of benchmark wins. Figure~\ref{fig:broad_reasoning_categories} groups these results by task type without aggregating scores within categories.

Figure~\ref{fig:broad_code_instruction_categories} shows code generation and instruction following separately. LLM-Debate leads on HumanEval and MATH; ChatEval leads on IFEval while scoring 4.44\% on AIME. No workflow leads on every benchmark. Appendix Figures~\ref{fig:appendix_accuracy_heatmap}--\ref{fig:appendix_delta_lollipops} show exploratory views of Run~1.

\paragraph{Case study: answer changes within one run.}
We compare the first Math Expert answer with the final output in a subset of records from one ChatEval GSM8K run. Among 371 tasks with a directly parseable numeric initial answer, 36 change from correct to incorrect and 6 change from incorrect to correct. Appendix~\ref{subsec:appendix_chateval_case_study} reports the transitions and examples, including a correct answer retained in role responses but lost before final output. This motivates checking answer preservation across the full workflow.

\subsection{GAIA Results and Backend Checks}
\label{subsec:gaia_results}
GAIA extends the evaluation to multi-step retrieval, file inspection, and tool use.
Table~\ref{tab:main_exp2_gaia} compares BenchAgent-compatible workflows with GPT-4.1 under the shared evaluation setup; all results in this subsection are single-run measurements.
\begin{table}[!ht]
    \centering
    \caption{
        \textbf{Single-run GAIA validation accuracy comparison within BenchAgent.} All workflows share the model, task set, tool registry, and evaluator. Overall pools correct answers over all 165 instances. Tok. reports mean recorded token usage per evaluated instance.
    }
    \label{tab:main_exp2_gaia}
    \setlength{\tabcolsep}{3pt}
    \small
    \renewcommand{\arraystretch}{1.2}
            \begin{tabular*}{\linewidth}{@{\extracolsep{\fill}}lrrrrr@{}}
            \toprule
 \textbf{Method}  &  \textbf{L1}  &  \textbf{L2}  &  \textbf{L3}  &  \textbf{Overall}  &  \textbf{Tok.} \\
            \midrule
            \multicolumn{6}{@{}l}{\emph{Single-Agent}} \\
            BenchAgent Core  &  58.49\%  &  30.23\%  &  19.23\%  &  37.58\%  &  459652.83 \\
            \midrule
            \multicolumn{6}{@{}l}{\emph{Evolving MAS}} \\
            EvoAgent  &  54.72\%  &  30.23\%  &  19.23\%  &  36.36\%  &  1573918.58 \\
            \multicolumn{6}{@{}l}{\emph{Fixed MAS}} \\
            LLM-Debate  &  52.83\%  &  33.72\%  &  15.38\%  &  36.97\%  &  414812.44 \\
            CAMEL  &  54.72\%  &  33.72\%  &  26.92\%  &  39.39\%  &  468587.39 \\
            AutoGen  &  56.60\%  &  30.23\%  &  11.54\%  &  35.76\%  &  383190.65 \\
            Jarvis  &  \textbf{66.04\%}  &  43.02\%  &  19.23\%  &  46.67\%  &  332285.96 \\
            ChatEval  &  49.06\%  &  34.88\%  &  26.92\%  &  38.18\%  &  1886517.67 \\
            \bottomrule
        \end{tabular*}
    \renewcommand{\arraystretch}{1.0}
\end{table}

Jarvis has the highest overall accuracy at 46.67\%, compared with 37.58\% for Core.
EvoAgent's positive broad-suite lift does not carry over to this GAIA run, where it reaches 36.36\%.
Table~\ref{tab:main_qwen32b_gaia_generalization} gives the Qwen3-32B workflow comparison, and Appendix Table~\ref{tab:appendix_glm5_gaia_generalization} reports a separate GLM-5 calibration of Core.

An additional analysis pairs retained Jarvis and ChatEval responses on 163 GAIA tasks. ChatEval records more tokens on 162 tasks, with 5.72 times Jarvis's total recorded usage; 33 tasks are correct only for Jarvis and 18 only for ChatEval. These historical batches provide descriptive evidence of task-dependent returns to expenditure, with execution-limit signals and answer-preservation examples in Appendix~\ref{subsec:appendix_paired_gaia}. This subset does not change the 165-task comparison in Table~\ref{tab:main_exp2_gaia}.

\begin{table}[!ht]
    \centering
    \caption{
        \textbf{Single-run GAIA validation results with Qwen3-32B.} This backend check uses the reported model settings and the same metrics as Table~\ref{tab:main_exp2_gaia}. Tok. reports mean recorded token usage per evaluated instance.
    }
    \label{tab:main_qwen32b_gaia_generalization}
    \setlength{\tabcolsep}{3pt}
    \small
    \renewcommand{\arraystretch}{1.2}
            \begin{tabular*}{\linewidth}{@{\extracolsep{\fill}}lrrrrr@{}}
            \toprule
  \textbf{Method}      &   \textbf{L1}                 &   \textbf{L2}   &   \textbf{L3}   &   \textbf{Overall}   &   \textbf{Tok.}  \\
            \midrule
            \multicolumn{6}{@{}l}{\emph{Single-Agent}} \\
            BenchAgent Core  &   47.17\%   &   24.42\%   &   19.23\%   &   30.91\%   &   269178.27  \\
            \midrule
            \multicolumn{6}{@{}l}{\emph{Evolving MAS}} \\
            EvoAgent  &   26.42\%   &   23.26\%   &   23.08\%   &   24.24\%   &   925891.35  \\
            \multicolumn{6}{@{}l}{\emph{Fixed MAS}} \\
            LLM-Debate  &   32.08\%   &   12.79\%   &   3.85\%   &   17.58\%   &   265544.08  \\
            CAMEL  &   39.62\%   &   23.26\%   &   3.85\%   &   25.45\%   &   93501.17  \\
            AutoGen  &   22.64\%   &   16.28\%   &   19.23\%   &   18.79\%   &   310282.87  \\
            Jarvis  &   39.62\%   &   18.60\%   &   23.08\%   &   26.06\%   &   213783.77  \\
            ChatEval  &   28.30\%   &   17.44\%   &   7.69\%   &   19.39\%   &   479251.38  \\
            \bottomrule
        \end{tabular*}
    \begin{minipage}{0.96\linewidth}
        \small
        With Qwen3-32B, all evaluated fixed and evolving MAS workflows trail BenchAgent Core in overall GAIA accuracy. This check covers one run on a second backend.
    \end{minipage}
    \renewcommand{\arraystretch}{1.0}
\end{table}

\subsection{Supplementary Case Study: Claude Code on GAIA}
\label{subsec:cc_case_study}
We also evaluate a dynamic Claude Code workflow on the same 165 GAIA validation instances, using GPT-4.1.
The CC-workflow and BenchAgent reference configurations share the GPT-4.1 backend family, task inputs, final-answer format, evaluator, and tool-capability classes. The CC-workflow has a different runtime, batching regime, and execution limits, so we report its behavior and performance as a separate observational case study.
Appendix Table~\ref{tab:appendix_gaia_protocol_summary} documents these settings.

The CC-workflow answers 110 of 165 tasks correctly (66.67\%), compared with 77 (46.67\%) for Jarvis, the highest-scoring BenchAgent configuration in this GAIA run, and records fewer tokens per instance (Table~\ref{tab:cc_case_study}).
Retained trajectories show agent creation, inter-agent messages, evidence artifacts, and verification events (Appendix~\ref{subsec:appendix_runtime_evidence}).
Figure~\ref{fig:gaia_generated_team_failure} in the appendix illustrates one same-instance contrast.
The reference rows contextualize the observed accuracy--token operating point. Its runtime and execution conditions limit comparison with BenchAgent, and the retained records support descriptions of execution behavior rather than attribution of gains to particular runtime features.

\begin{table}[H]
    \centering
    \caption{Supplementary single-run GAIA case study. Overall accuracy pools correct answers over 165 tasks. Token means summarize recorded LLM usage per evaluated instance. BenchAgent rows reproduce reference results from Table~\ref{tab:main_exp2_gaia} to provide context; CC uses a different runtime, batching regime, and execution limits.}
    \label{tab:cc_case_study}
    \small
    \setlength{\tabcolsep}{4pt}
    \begin{tabular}{@{}lrr@{}}
        \toprule
        Method & Overall Acc. & Mean tokens \\
        \midrule
        BenchAgent Core & 37.58\% & 459652.83 \\
        Jarvis & 46.67\% & 332285.96 \\
        \midrule
        CC-workflow & 66.67\% & 52984.69 \\
        \bottomrule
    \end{tabular}
\end{table}

\section{Discussion}
The choice of workflow depends on which tasks and costs matter for an evaluation.
EvoAgent improves pooled accuracy, whereas LLM-Debate and ChatEval lead on particular benchmarks despite lower pooled scores.
CAMEL and Jarvis use fewer tokens but also have lower accuracy, so a lower pooled score alone says little about a workflow's token cost.
The pattern also differs by backend: all evaluated MAS trail Core in the Qwen3-32B GAIA run, while several exceed it with GPT-4.1.
The response analyses add concrete optimization targets to these performance profiles: examining resource allocation when repeated calls reach execution limits, and checking whether correct role answers survive aggregation and extraction. These are hypotheses for subsequent controlled tests. The shared protocol allows resulting workflow changes to be compared on the same tasks.

\paragraph{Future work.}
Varying the shared Core step limits can characterize each workflow's accuracy--token profile.
Component ablations can separate the contributions of prompt-scaffold search, aggregation, delegation, verification, and context handling from the effects of additional calls and differing execution policies.
Controlled studies of the external runtime would require matched tools and execution conditions; repeated GAIA runs could assess the stability of the reported snapshot.
Testing other controllers, backends, and open-ended interactive tasks would help establish how far these patterns generalize.

\section{Conclusion}
BenchAgent provides a common basis for comparing the accuracy and token expenditure of single- and multi-agent workflows. Across seven configurations, ten benchmarks, and three runs with GPT-4.1, the results reveal task-specific strengths, regressions, and cost trade-offs. Analyses of retained responses complement this broad evidence with concrete questions about resource allocation and answer preservation. Together, the framework and results support the evaluation of subsequent multi-agent workflow improvements under the same protocol.

\section{Limitations}
\label{sec:limitations}
The findings apply to the evaluated implementations and task sets. We measure all workflow lifts relative to BenchAgent Core, using normalized re-instantiations of published MAS designs within the shared execution setting.
Shared Core step limits constrain each execution, while workflows induce different call counts and realized token use. The comparisons characterize complete configurations within the normalized environment; agent-count and individual component effects remain outside this measurement target.

The broad-suite measurements cover three runs on one fixed task set per benchmark and do not measure variation across task samples. GAIA and auxiliary backend checks remain single-run measurements.
Uneven coverage of process records limits generalization and causal interpretation of the case analyses.
The external CC case study describes one configuration under its own runtime and execution conditions. Its traces expose only part of its internal execution, limiting reproducibility and interpretation of individual components.
The benchmark suite covers verifiable tasks and does not represent all open-ended interactive deployments.

%% file: resources/appendix.tex
\section{Evaluation Protocol and Supplementary Evidence}

This appendix describes dataset sampling, evaluation settings, workflow implementations, auxiliary results, and example trajectories.

\subsection{Protocol and Configuration Normalization}
\label{subsec:appendix_protocol_fidelity}

\paragraph{Dataset sources.}
The ten broad benchmarks use processed evaluation subsets distributed in \href{https://huggingface.co/datasets/Brian-Wang117/MASArena-Dataset/tree/468540bdc34bc0f8c7ed52185bcdc8281287b6c3}{\texttt{Brian-Wang117/}\allowbreak\texttt{MASArena-Dataset}}, pinned to revision \texttt{468540bdc34bc0f8c7ed52185bcdc8281287b6c3}. Table~\ref{tab:appendix_dataset_sampling} lists their sources and evaluated sizes. AFlow-derived subsets follow the data distributed by the \href{https://github.com/FoundationAgents/AFlow/blob/main/data/download_data.py}{AFlow repository}. GAIA uses the 2023 validation split from \href{https://huggingface.co/datasets/gaia-benchmark/GAIA}{\texttt{gaia-benchmark/GAIA}}.

\paragraph{Sampling.}
For processed source files containing more than 400 instances, we uniformly sample 400 without replacement using Python's \texttt{random.sample} with seed 42. This applies to MATH, GSM8K, BBH, HotpotQA, and IFEval. Files with at most 400 instances are evaluated in full. The ten broad benchmarks total 3,302 instances per run; GAIA uses all 165 validation instances, with level sizes $(53,86,26)$.

\begingroup
\small
\setlength{\tabcolsep}{5pt}
\renewcommand{\arraystretch}{1.10}
\setlength{\LTcapwidth}{\linewidth}
\begin{longtable}{@{}>{\raggedright\arraybackslash}p{\dimexpr 0.14286\linewidth - 0.85714\tabcolsep\relax}>{\raggedright\arraybackslash}p{\dimexpr 0.53846\linewidth - 3.23077\tabcolsep\relax}>{\raggedright\arraybackslash}p{\dimexpr 0.10989\linewidth - 0.65934\tabcolsep\relax}>{\raggedright\arraybackslash}p{\dimexpr 0.20879\linewidth - 1.25275\tabcolsep\relax}@{}}
\caption{Dataset sources and sampling from the pinned MASArena snapshot. ``All'' denotes the complete processed file; larger files are sampled with seed 42.}
    \label{tab:appendix_dataset_sampling} \\
\toprule
        \textbf{Dataset} & \textbf{Source / version} & \textbf{Eval. $N$} & \textbf{Sampling} \\
        \midrule
\endfirsthead
\caption[]{\textit{Continued.}} \\
\toprule
        \textbf{Dataset} & \textbf{Source / version} & \textbf{Eval. $N$} & \textbf{Sampling} \\
        \midrule
\endhead
\midrule
\multicolumn{4}{r}{\footnotesize Continued on next page} \\
\endfoot
\bottomrule
\endlastfoot
MATH & AFlow-processed MATH Level-5 subset. & 400 & Seed 42 \\
        AIME & AIME 2025 I + II, using the MASArena text version. & 30 & All \\
        GSM8K & \href{https://huggingface.co/datasets/openai/gsm8k}{\texttt{openai/gsm8k}}, main/test; AFlow subset. & 400 & Seed 42 \\
        DROP & AFlow-processed DROP, first 400 records of its 800-instance test file. & 400 & All \\
        BBH & \href{https://huggingface.co/datasets/lukaemon/bbh}{\texttt{lukaemon/bbh}}, test; MASArena subset. & 400 & Seed 42 \\
        MMLU-Pro & \href{https://huggingface.co/datasets/TIGER-Lab/MMLU-Pro/tree/475d58ba0cc18a15fd5d4221f41919199e692331}{\texttt{TIGER-Lab/MMLU-Pro}}, test at revision \texttt{475d58ba}; MASArena subset. & 400 & All \\
        HumanEval & \href{https://huggingface.co/datasets/openai/openai_humaneval}{\texttt{openai/openai\_humaneval}}, test; AFlow subset. & 131 & All \\
        MBPP & AFlow-processed sanitized MBPP. & 341 & All \\
        HotpotQA & \href{https://huggingface.co/datasets/hotpotqa/hotpot_qa}{\texttt{hotpotqa/hotpot\_qa}}, distractor/validation; AFlow subset. & 400 & Seed 42 \\
        IFEval & IFEval version distributed in the pinned MASArena snapshot. & 400 & Seed 42 \\
        GAIA & \href{https://huggingface.co/datasets/gaia-benchmark/GAIA}{\texttt{gaia-benchmark/GAIA}}, 2023 validation. & 165 & All \\
\end{longtable}
\endgroup

\input{resources/verified_counts.tex}

\input{resources/scoring_protocol.tex}

\begingroup
\small
\setlength{\tabcolsep}{5pt}
\renewcommand{\arraystretch}{1.10}
\setlength{\LTcapwidth}{\linewidth}
\begin{longtable}{@{}>{\raggedright\arraybackslash}p{\dimexpr 0.43478\linewidth - 1.73913\tabcolsep\relax}>{\raggedright\arraybackslash}p{\dimexpr 0.19565\linewidth - 0.78261\tabcolsep\relax}>{\raggedright\arraybackslash}p{\dimexpr 0.36957\linewidth - 1.47826\tabcolsep\relax}@{}}
\caption{Tool regimes for the reported experiments. All workflows also share the final-answer interface.}
    \label{tab:appendix_tool_regimes} \\
\toprule
        \textbf{Dataset(s)} & \textbf{Experiment} & \textbf{Tool regime} \\
        \midrule
\endfirsthead
\caption[]{\textit{Continued.}} \\
\toprule
        \textbf{Dataset(s)} & \textbf{Experiment} & \textbf{Tool regime} \\
        \midrule
\endhead
\midrule
\multicolumn{3}{r}{\footnotesize Continued on next page} \\
\endfoot
\bottomrule
\endlastfoot
MATH, AIME, GSM8K, DROP, BBH, MMLU-Pro, HumanEval, MBPP, IFEval & Broad benchmark & \texttt{python\_interpreter} only \\
        HotpotQA & Broad benchmark & Expanded full BenchAgent tool registry \\
        GAIA validation (BenchAgent-based systems) & Runtime-workflow study & Expanded full BenchAgent tool registry \\
        GAIA validation (CC-workflow) & Runtime-workflow study & Corresponding GAIA-relevant tool-capability classes \\
\end{longtable}
\endgroup

\paragraph{Reported model configuration.}
The reported run settings are \texttt{MAX\_TOKEN\_SIZE}=8192, \texttt{temperature}=0.2, and \texttt{top\_p}=1.0 unless a role-specific setting is stated. Tables~\ref{tab:appendix_benchagent_core_prompt}--\ref{tab:appendix_evoagent_contract} describe the implementations used in the reported experiments.

\paragraph{Shared execution limits.}
The baseline controller and Core executors invoked by MAS use the same manager \texttt{max\_steps}=15 and search \texttt{max\_steps}=10. Workflow-specific dialogue rounds, candidate populations, and aggregation determine how these executions are organized. Token use records the resulting expenditure under these limits.

\begingroup
\small
\setlength{\tabcolsep}{4pt}
\renewcommand{\arraystretch}{1.10}
\setlength{\LTcapwidth}{\linewidth}
\begin{longtable}{@{}>{\raggedright\arraybackslash}p{\dimexpr 0.20879\linewidth - 0.83516\tabcolsep\relax}>{\raggedright\arraybackslash}p{\dimexpr 0.40659\linewidth - 1.62637\tabcolsep\relax}>{\raggedright\arraybackslash}p{\dimexpr 0.38462\linewidth - 1.53846\tabcolsep\relax}@{}}
\caption{Shared interfaces and workflow-dependent components in the BenchAgent comparison.}
    \label{tab:appendix_shared_interfaces} \\
\toprule
        \textbf{Interface} & \textbf{Shared component} & \textbf{Workflow-dependent component} \\
        \midrule
\endfirsthead
\caption[]{\textit{Continued.}} \\
\toprule
        \textbf{Interface} & \textbf{Shared component} & \textbf{Workflow-dependent component} \\
        \midrule
\endhead
\midrule
\multicolumn{3}{r}{\footnotesize Continued on next page} \\
\endfoot
\bottomrule
\endlastfoot
Benchmark input & Evaluated task set and normalized instance schema & Role-specific use of the instance context \\
        Model backend & Configured model and common model-call settings & Number and sequence of calls \\
        Tool access & Benchmark-specific tool interface and available capabilities & Tool selection and role-level access \\
        Answer evaluation & Final-answer extraction and benchmark evaluator & Proposal generation and aggregation \\
        Token accounting & Recorded input and output usage & Realized per-instance consumption \\
        Trajectory logging & Common message, tool-call, and usage records & Role interactions and execution sequence \\
\end{longtable}
\endgroup

We evaluate additional model backends on GAIA. Table~\ref{tab:main_qwen32b_gaia_generalization} compares all BenchAgent workflows using Qwen3-32B, and Table~\ref{tab:appendix_glm5_gaia_generalization} reports Core results using GLM-5.

\begingroup
\small
\setlength{\tabcolsep}{4pt}
\setlength{\LTcapwidth}{\linewidth}
\renewcommand{\arraystretch}{1.10}
\begin{longtable}{l l ccc}
\caption{\small
        \textbf{BenchAgent Core on GAIA with GLM-5.} Cells report accuracy / mean tokens per instance.
    }
    \label{tab:appendix_glm5_gaia_generalization} \\
\toprule
            \textbf{Paradigm} & \textbf{Method} & \textbf{GAIA-L1} & \textbf{GAIA-L2} & \textbf{GAIA-L3} \\
            \midrule
\endfirsthead
\caption[]{\textit{Continued.}} \\
\toprule
            \textbf{Paradigm} & \textbf{Method} & \textbf{GAIA-L1} & \textbf{GAIA-L2} & \textbf{GAIA-L3} \\
            \midrule
\endhead
\midrule
\multicolumn{5}{r}{\footnotesize Continued on next page} \\
\endfoot
\bottomrule
\endlastfoot
Single-Agent & BenchAgent & 81.13\% / 268743.54 & 63.95\% / 418454.21 & 38.46\% / 3375974 \\
\end{longtable}
\endgroup

All evaluated MAS score below Core in the Qwen3-32B GAIA check. Replicating the ten-benchmark suite across backends remains future work.

\begingroup
\small
\setlength{\tabcolsep}{4pt}
\renewcommand{\arraystretch}{1.10}
\setlength{\LTcapwidth}{\linewidth}
\begin{longtable}{@{}>{\raggedright\arraybackslash}p{\dimexpr 0.21739\linewidth - 0.86957\tabcolsep\relax}>{\raggedright\arraybackslash}p{\dimexpr 0.46739\linewidth - 1.86957\tabcolsep\relax}>{\raggedright\arraybackslash}p{\dimexpr 0.31522\linewidth - 1.26087\tabcolsep\relax}@{}}
\caption{Workflow categories used in the paper.}
    \label{tab:appendix_workflow_taxonomy} \\
\toprule
        \textbf{Category} & \textbf{Definition} & \textbf{Systems in this paper} \\
        \midrule
\endfirsthead
\caption[]{\textit{Continued.}} \\
\toprule
        \textbf{Category} & \textbf{Definition} & \textbf{Systems in this paper} \\
        \midrule
\endhead
\midrule
\multicolumn{3}{r}{\footnotesize Continued on next page} \\
\endfoot
\bottomrule
\endlastfoot
Single-agent & One task-level controller maintains context, calls shared tools (including a search helper), and emits the final answer. & BenchAgent Core \\
        Fixed MAS & Role types and stage-level communication rules are predefined; executor instantiation and assignment may depend on the task. & CAMEL, AutoGen instantiation, Jarvis/HuggingGPT-style wrapper, ChatEval, LLM-Debate \\
        Evolving MAS & Search over a predefined configuration family; the tested implementation evolves agent prompts. & EvoAgent \\
        Runtime-generated workflow & The controller can create task-specific agents, state artifacts, tool scopes, and verification branches during execution. & CC-workflow with a GPT-4.1 backend \\
\end{longtable}
\endgroup

\begingroup
\small
\setlength{\tabcolsep}{3.5pt}
\renewcommand{\arraystretch}{1.10}
\setlength{\LTcapwidth}{\linewidth}
\begin{longtable}{@{}>{\raggedright\arraybackslash}p{\dimexpr 0.24731\linewidth - 0.98925\tabcolsep\relax}>{\raggedright\arraybackslash}p{\dimexpr 0.37634\linewidth - 1.50538\tabcolsep\relax}>{\raggedright\arraybackslash}p{\dimexpr 0.37634\linewidth - 1.50538\tabcolsep\relax}@{}}
\caption{GAIA evaluation settings for BenchAgent and the CC case study.}
    \label{tab:appendix_gaia_protocol_summary} \\
\toprule
        \textbf{Field} & \textbf{BenchAgent-Based Systems} & \textbf{Claude Code Configuration} \\
        \midrule
\endfirsthead
\caption[]{\textit{Continued.}} \\
\toprule
        \textbf{Field} & \textbf{BenchAgent-Based Systems} & \textbf{Claude Code Configuration} \\
        \midrule
\endhead
\midrule
\multicolumn{3}{r}{\footnotesize Continued on next page} \\
\endfoot
\bottomrule
\endlastfoot
Backbone and provider & Official OpenAI \texttt{gpt-4.1-2025-04-14} snapshot via the provider family used in the main experiments & OpenAI \texttt{gpt-4.1-2025-04-14} through LiteLLM request translation and an OpenAI-compatible provider \\
        Model-call parameters & Reported run settings: \texttt{MAX\_TOKEN\_SIZE}=8192; \texttt{temperature}=0.2; \texttt{top\_p}=1.0 & Reported \texttt{MAX\_TOKEN\_SIZE}=8192; retained records show \texttt{temperature}=0.2 and \texttt{top\_p}=1.0 for OpenAI-compatible calls \\
        Invocation and version & BenchAgent runner scripts under the shared evaluation harness & Claude Code VS Code plugin, version 2.1.68 \\
        Reporting protocol & Final-answer accuracy & Final-answer accuracy from extracted outputs \\
        Evaluation split & GAIA validation, Levels 1--3; $(N_1,N_2,N_3)=(53,86,26)$ & GAIA validation, Levels 1--3; $(N_1,N_2,N_3)=(53,86,26)$ \\
        Execution limits & manager \texttt{max\_steps}\allowbreak=15; search \texttt{max\_steps}=10; 600s API timeout; 300s default MCP/server timeout & Prompt-level batching and a 60s shell-command timeout; other controller and provider limits are unspecified. Resource budgets are not matched. \\
        Tool and server configuration & expanded full BenchAgent tool registry for GAIA & Claude Code built-in tools plus Agent Teams orchestration; no MCP or external servers were enabled \\
        Permission and sandbox surface & task-level sandbox, tool allowlists, and MCP/server timeouts controlled by BenchAgent & Unattended execution; prompts restrict writes to \texttt{outputs/**} and \texttt{logs/**}; settings deny secrets and destructive deletion \\
        Request retry and stopping & \texttt{max\_retries}=3 for API requests; stop on final answer, step budget, or runtime error; evaluator verdicts do not trigger task-level reruns & Repeated/resumed calls contribute to recorded usage; final answers are extracted from \texttt{outputs/task\_id.txt}, with existing outputs skipped \\
        Token accounting & Prompt/completion metadata aggregated across agent messages & Prompt/completion metadata for lead-agent, subagent, tool-emitting, and auxiliary calls \\
        Process evidence & BenchAgent message and tool logs & Team-creation, agent, and role-message events summarized in Section~\ref{subsec:appendix_runtime_evidence} \\
\end{longtable}
\endgroup

\paragraph{Claude Code configuration notes.}
The CC-workflow enables experimental Agent Teams with Planner, Solver, Verifier, and Writer roles. Table~\ref{tab:appendix_claude_prompt_contract} gives its prompt and output contract.

\paragraph{Model routing.}
Role-spawn records retain the client label \texttt{model: sonnet}. LiteLLM translates the request body to invoke \texttt{gpt-4.1-2025-04-14}; manual checks of the API provider confirmed this routing. The retained role labels therefore describe the client configuration, while the provider calls use GPT-4.1.

\begingroup
\small
\setlength{\tabcolsep}{3pt}
\renewcommand{\arraystretch}{1.10}
\setlength{\LTcapwidth}{\linewidth}
\begin{longtable}{@{}>{\raggedright\arraybackslash}p{\dimexpr 0.18085\linewidth - 1.08511\tabcolsep\relax}>{\raggedright\arraybackslash}p{\dimexpr 0.26596\linewidth - 1.59574\tabcolsep\relax}>{\raggedright\arraybackslash}p{\dimexpr 0.28723\linewidth - 1.72340\tabcolsep\relax}>{\raggedright\arraybackslash}p{\dimexpr 0.26596\linewidth - 1.59574\tabcolsep\relax}@{}}
\caption{GAIA tool capabilities and logged events. BenchAgent counts come from \texttt{log.rar} and the selected Jarvis/ChatEval batches in Tables~\ref{tab:appendix_benchagent_gaia_log_counts}--\ref{tab:appendix_selected_gaia_response_log_stats}.}
\label{tab:appendix_gaia_tool_alignment} \\
\toprule
\textbf{Capability} & \textbf{BenchAgent Configured Surface} & \textbf{BenchAgent Retained Evidence} & \textbf{Claude Code Evidence} \\
\midrule
\endfirsthead
\caption[]{\textit{Continued.}} \\
\toprule
\textbf{Capability} & \textbf{BenchAgent Configured Surface} & \textbf{BenchAgent Retained Evidence} & \textbf{Claude Code Evidence} \\
\midrule
\endhead
\midrule
\multicolumn{4}{r}{\footnotesize Continued on next page} \\
\endfoot
\bottomrule
\endlastfoot
Web search & Web-search and retrieval tools in the GAIA registry. & \texttt{log.rar}: 11{,}050 \texttt{Request URL} events (17.40/inst.). Selected ChatEval: 2{,}338 retrieval-step calls and 4{,}449 executed retrieval records. & Schema exposes \texttt{WebSearch}; retained structured events include 3 calls. \\
Web page fetching / browsing & Web, crawler, browser, and page-reader tools. & Selected ChatEval: 4{,}925 executed external-tool records (29.85/inst.); browser/crawl/archive tools appear among top executed tools. & Web-fetch capability is visible. \\
Local file discovery & Benchmark-visible files and file readers. & Discovery is not separated from benchmark file injection in retained BenchAgent logs. & 2 retained \texttt{Glob} calls. \\
Local file reading & Text, Markdown, document, and file readers. & Selected ChatEval: 231 executed file-tool records (1.40/inst.); Jarvis and \texttt{log.rar} do not serialize typed file reads consistently. & 29 retained \texttt{Read} calls; additional inspection often goes through \texttt{Bash}. \\
Shell and code execution & Python/interpreter-style execution through the shared runtime. & \texttt{log.rar}: 159 \texttt{Code Agent Step} and 59 \texttt{Executing code} markers. Selected ChatEval: 3{,}615 structured code-tool calls. & 207 retained \texttt{Bash} events. \\
Structured tables & CSV and sheet-style extraction tools. & CSV/sheet inspection is included in selected ChatEval file-tool records. & File tools, shell commands, or Python. \\
Document / text parsing & Text and document parsing paths. & Text extraction and Markdown conversion appear among selected ChatEval top tools. & Direct file reads and shell-based parsing. \\
Writing artifacts & Answer, trajectory, and usage serialization. & Selected files retain 328 responses (165 Jarvis and 163 ChatEval) for 330 evaluation instances and serialize messages, usage, and final answers; \texttt{tool\_calls} are absent. & 65 retained \texttt{Write} events; prompt restricts writes to \texttt{outputs/} and \texttt{logs/}. \\
        Team orchestration & Fixed, dynamic, or debate-style MAS wrappers. & Selected responses: ChatEval has 1{,}878 model calls (11.38/inst.); Jarvis has 165 (1.00/inst.). & 7 \texttt{TeamCreate}, 31 \texttt{Agent}, and 35 \texttt{SendMessage} events. \\
Task/state management & Manager state, step budgets, messages, tool calls, and evaluator outputs. & Selected responses: 3{,}756 ChatEval messages (22.76/inst.) and 330 Jarvis messages (2.00/inst.). & Role-scoped messages, written artifacts, final-answer files, and \texttt{TodoWrite} schema are visible. \\
External servers & Optional MCP/server-backed registered tools. & Retained logs do not separate MCP/server-backed calls from ordinary registered tool calls. & No MCP or external servers were enabled. \\
\end{longtable}
\endgroup

\begingroup
\small
\setlength{\tabcolsep}{4pt}
\setlength{\LTcapwidth}{\linewidth}
\renewcommand{\arraystretch}{1.10}
\begin{longtable}{lcccccc}
\caption{GAIA event counts from \texttt{log.rar}; parentheses give means per logged instance. Counts cover available log markers, not all internal tool activity: ``Code-agent steps'' counts \texttt{Code Agent Step}; ``Direct external'' counts \texttt{Calling tool:} excluding \texttt{final\_answer}; ``Wiki/API req.'' counts \texttt{Request URL:}; ``Code exec.'' counts \texttt{Executing code:}; ``Tool errors'' counts tool-execution errors.}
    \label{tab:appendix_benchagent_gaia_log_counts} \\
\toprule
        \textbf{Method} & \shortstack{\textbf{Retained}\\\textbf{instances}} & \shortstack{\textbf{Code-agent}\\\textbf{steps}} & \shortstack{\textbf{Direct}\\\textbf{external}} & \shortstack{\textbf{Wiki/API}\\\textbf{req.}} & \textbf{Code exec.} & \textbf{Tool errors} \\
        \midrule
\endfirsthead
\caption[]{\textit{Continued.}} \\
\toprule
        \textbf{Method} & \shortstack{\textbf{Retained}\\\textbf{instances}} & \shortstack{\textbf{Code-agent}\\\textbf{steps}} & \shortstack{\textbf{Direct}\\\textbf{external}} & \shortstack{\textbf{Wiki/API}\\\textbf{req.}} & \textbf{Code exec.} & \textbf{Tool errors} \\
        \midrule
\endhead
\midrule
\multicolumn{7}{r}{\footnotesize Continued on next page} \\
\endfoot
\bottomrule
\endlastfoot
Single Agent & 127 & 45 (0.35) & 116 (0.91) & 2301 (18.12) & 33 (0.26) & 7 (0.06) \\
        EvoAgent & 127 & 50 (0.39) & 0 (0.00) & 2429 (19.13) & 0 (0.00) & 15 (0.12) \\
        LLM-Debate & 127 & 15 (0.12) & 0 (0.00) & 1987 (15.65) & 2 (0.02) & 9 (0.07) \\
        CAMEL & 127 & 21 (0.17) & 10 (0.08) & 2433 (19.16) & 7 (0.06) & 7 (0.06) \\
        AutoGen & 127 & 28 (0.22) & 18 (0.14) & 1900 (14.96) & 17 (0.13) & 10 (0.08) \\
\end{longtable}
\endgroup

\begingroup
\small
\setlength{\tabcolsep}{4pt}
\setlength{\LTcapwidth}{\linewidth}
\renewcommand{\arraystretch}{1.10}
\begin{longtable}{lcccccc}
\caption{GAIA statistics from six selected Jarvis/ChatEval result files. The evaluation denominator remains 165 per method (330 total); retained matched responses number 165 for Jarvis and 163 for ChatEval (328 total). Parentheses normalize counts by the evaluation denominator of 165, not by retained-response count. Model calls come from response files; tool events come from logs. ``Step external'' counts \texttt{STEP\_LOG} calls; ``executed external'' counts \texttt{TOOL\_CALL} records; ``direct/code markers'' combines \texttt{Calling tool:} and \texttt{Executing code:}. Tool-call counts exclude \texttt{final\_answer}. Jarvis's structured tool events are not serialized.}
    \label{tab:appendix_selected_gaia_response_log_stats} \\
\toprule
        \textbf{Method} & \shortstack{\textbf{Evaluation}\\\textbf{$N$}} & \shortstack{\textbf{Matched}\\\textbf{responses}} & \shortstack{\textbf{Model}\\\textbf{calls}} & \shortstack{\textbf{Step}\\\textbf{external}} & \shortstack{\textbf{Executed}\\\textbf{external}} & \shortstack{\textbf{Direct/code}\\\textbf{markers}} \\
        \midrule
\endfirsthead
\caption[]{\textit{Continued.}} \\
\toprule
        \textbf{Method} & \shortstack{\textbf{Evaluation}\\\textbf{$N$}} & \shortstack{\textbf{Matched}\\\textbf{responses}} & \shortstack{\textbf{Model}\\\textbf{calls}} & \shortstack{\textbf{Step}\\\textbf{external}} & \shortstack{\textbf{Executed}\\\textbf{external}} & \shortstack{\textbf{Direct/code}\\\textbf{markers}} \\
        \midrule
\endhead
\midrule
\multicolumn{7}{r}{\footnotesize Continued on next page} \\
\endfoot
\bottomrule
\endlastfoot
Jarvis & 165 & 165 (1.00) & 165 (1.00) & \shortstack{0\\(not serialized)} & \shortstack{0\\(not serialized)} & 22 (0.13) \\
        ChatEval & 165 & 163 (0.99) & 1878 (11.38) & 5953 (36.08) & 4925 (29.85) & 1476 (8.95) \\
\end{longtable}
\endgroup

\begingroup
\small
\setlength{\tabcolsep}{4pt}
\renewcommand{\arraystretch}{1.10}
\setlength{\LTcapwidth}{\linewidth}
\begin{longtable}{@{}>{\raggedright\arraybackslash}p{\dimexpr 0.25000\linewidth - 0.50000\tabcolsep\relax}>{\raggedright\arraybackslash}p{\dimexpr 0.75000\linewidth - 1.50000\tabcolsep\relax}@{}}
\caption{Claude Code GAIA prompt and output contract.}
    \label{tab:appendix_claude_prompt_contract} \\
\toprule
        \textbf{Prompt Component} & \textbf{Retained Content / Operational Effect} \\
        \midrule
\endfirsthead
\caption[]{\textit{Continued.}} \\
\toprule
        \textbf{Prompt Component} & \textbf{Retained Content / Operational Effect} \\
        \midrule
\endhead
\midrule
\multicolumn{2}{r}{\footnotesize Continued on next page} \\
\endfoot
\bottomrule
\endlastfoot
Autonomy rule & The prompt instructs the system to operate autonomously as an Agent Team and continue without waiting for user confirmation unless a critical error occurs. \\
        Team roles & The retained run contains four named roles: Planner, Solver, Verifier, and Writer. Planner inspects the project and task split; Solver reasons over each task; Verifier checks the proposed answer; Writer emits final answer files and logs. \\
        Input-reading rule & The prompt asks the Planner to inspect JSONL split files using \texttt{head}, \texttt{sed}, and \texttt{tail}. \\
        Split and batching & Retained prompts include both a first-10-task batch contract and a GAIA validation contract over \texttt{split\_validation/gaia1\_2.jsonl}; existing output files are skipped rather than overwritten. \\
        Write scope & The prompt restricts writes to \texttt{outputs/} and \texttt{logs/} and forbids modifying code, task data, prompts, configurations, and benchmark assets. \\
        Command restrictions & Destructive file deletion and network-download commands are forbidden by prompt-level rules; the retained settings also deny secrets and destructive deletion patterns. \\
        Final extraction & Answer-only \texttt{outputs/<task\_id>.txt} files provide final answers; \texttt{logs/} contains per-task reasoning or evidence. \\
\end{longtable}
\endgroup

\paragraph{Claude Code token accounting.}
We count input and output tokens across lead-agent, subagent, delegation, and auxiliary calls, including unsuccessful attempts. Mean usage is the total divided by the number of evaluated tasks. Instructions, prompts, tool schemas, and tool results count as input tokens when supplied to the model. The deduplicated batch contains 128 subagent model calls, 206 calls that emitted tool calls, 236 calls with historical \texttt{tool\_use}/\texttt{tool\_result} context, and 22 title-generation calls.

\subsection{Prompt and Trajectory Excerpts}
\label{subsec:appendix_prompt_traj_examples}

Tables~\ref{tab:appendix_benchagent_core_prompt}--\ref{tab:appendix_evoagent_contract} describe Core and the MAS wrappers; Tables~\ref{tab:appendix_retained_trajectory_excerpt}--\ref{tab:appendix_additional_trajectory_excerpts} illustrate their execution alongside the CC case study. Long task text, paths, and tool schemas are shortened for readability.

\paragraph{Scope of process evidence.}
Process records have unequal coverage and logging granularity. The five-method GAIA logs retain event markers but lack consistent task identifiers; Jarvis and ChatEval response exports provide answer snapshots, while selected CC records expose tool and message histories. These sources do not form a complete, task-aligned trajectory set across all seven BenchAgent workflows. We use them for descriptive process summaries and illustrative cases. The paired answer-transition analysis is restricted to 371 GSM8K tasks from one ChatEval run with parseable first-role answers and matched final scores; it does not estimate a suite-wide failure rate or isolate a coordination mechanism.

\begingroup
\small
\setlength{\tabcolsep}{4pt}
\renewcommand{\arraystretch}{1.10}
\setlength{\LTcapwidth}{\linewidth}
\begin{longtable}{@{}>{\raggedright\arraybackslash}p{\dimexpr 0.21739\linewidth - 0.86957\tabcolsep\relax}>{\raggedright\arraybackslash}p{\dimexpr 0.31522\linewidth - 1.26087\tabcolsep\relax}>{\raggedright\arraybackslash}p{\dimexpr 0.46739\linewidth - 1.86957\tabcolsep\relax}@{}}
\caption{BenchAgent Core prompt and execution contract, quoted or paraphrased from its prompt templates and runner.}
    \label{tab:appendix_benchagent_core_prompt} \\
\toprule
        \textbf{Prompt / code field} & \textbf{Retained wording or behavior} & \textbf{Evaluation role} \\
        \midrule
\endfirsthead
\caption[]{\textit{Continued.}} \\
\toprule
        \textbf{Prompt / code field} & \textbf{Retained wording or behavior} & \textbf{Evaluation role} \\
        \midrule
\endhead
\midrule
\multicolumn{3}{r}{\footnotesize Continued on next page} \\
\endfoot
\bottomrule
\endlastfoot
Task wrapper & ``Answer this question correctly. You have all the tools needed to find the right answer.'' & Gives every benchmark instance the same task-facing wrapper before workflow-specific execution begins. \\
        Tool priority & ``Always prioritize using available tools over writing custom code.'' & Encourages use of registered tools. \\
        File-access rule & ``If the task involves files or attachments, you MUST use appropriate tools to access them.'' & Encourages file access through the logged tool interface. \\
        Search delegation & The manager can call \texttt{search\_agent("Your detailed request")} for web research and online documents. & Separates manager reasoning from delegated web research while preserving a recorded handoff. \\
        Direct answer contract & The manager must call \texttt{final\_answer("Your answer")} and keep final answers concise. & Normalizes final-answer extraction across benchmarks and workflows. \\
        Runtime limits & The manager uses \texttt{max\_steps}\allowbreak=15 and the search agent uses \texttt{search\_max\_steps}=10 unless configured otherwise. & Makes step budgets explicit for the BenchAgent workflow comparison. \\
        Verification pressure & The task wrapper says failures or \texttt{None found} are not acceptable and asks the agent to verify when needed. & Encourages evidence checking without changing the final-answer-only evaluator. \\
        Trace serialization & Returned records include message history, final answer, manager steps, search-agent steps, and usage metadata when available. & Provides the process evidence used in the main text and appendix diagnostics. \\
\end{longtable}
\endgroup

\begingroup
\small
\setlength{\tabcolsep}{3.5pt}
\renewcommand{\arraystretch}{1.10}
\setlength{\LTcapwidth}{\linewidth}
\begin{longtable}{@{}>{\raggedright\arraybackslash}p{\dimexpr 0.10753\linewidth - 0.86022\tabcolsep\relax}>{\raggedright\arraybackslash}p{\dimexpr 0.15054\linewidth - 1.20430\tabcolsep\relax}>{\raggedright\arraybackslash}p{\dimexpr 0.32258\linewidth - 2.58065\tabcolsep\relax}>{\raggedright\arraybackslash}p{\dimexpr 0.27957\linewidth - 2.23656\tabcolsep\relax}>{\raggedright\arraybackslash}p{\dimexpr 0.13978\linewidth - 1.11828\tabcolsep\relax}@{}}
\caption{Roles, instructions, and execution procedures for five fixed MAS wrappers. ``LLM-only'' denotes no external tools beyond \texttt{final\_answer}.}
    \label{tab:appendix_reproduced_mas_contracts} \\
\toprule
        \textbf{System} & \textbf{Agent instances} & \textbf{Role-level instruction content} & \textbf{Communication and termination} & \textbf{Tools and limits} \\
        \midrule
\endfirsthead
\caption[]{\textit{Continued.}} \\
\toprule
        \textbf{System} & \textbf{Agent instances} & \textbf{Role-level instruction content} & \textbf{Communication and termination} & \textbf{Tools and limits} \\
        \midrule
\endhead
\midrule
\multicolumn{5}{r}{\footnotesize Continued on next page} \\
\endfoot
\bottomrule
\endlastfoot
Jarvis & LLM controller + task executors & \textbf{(1) Task Planning:} the controller decomposes the request into tasks. \textbf{(2) Model Selection:} instantiate an executor for each task. \textbf{(3) Task Execution:} invoke the executors' models to perform the assigned tasks. \textbf{(4) Response Generation:} the controller aggregates execution results into the final response. & Controller dispatches planned tasks to executors and receives their results for aggregation; \texttt{final\_answer} terminates the workflow. & Shared model backend and benchmark-specific tool regime; manager \texttt{max\_steps}\allowbreak=15 \\
        \addlinespace
        ChatEval & Math Expert; Logic Expert; Critical Thinking Expert; ResultExtractor & \textbf{Math Expert:} analyze mathematically; verify with Python; use search for current data and file tools for attachments. \textbf{Logic Expert:} analyze logical structure and implicit conditions; be concise. \textbf{Critical Thinking Expert:} analyze multiple angles and identify traps. Logic and Critical prompts ask the role to avoid tools. & 2 rounds of sequential Math, Logic, and Critical contributions with shared history and no consensus-based early stop. ResultExtractor produces the final answer. & BenchAgent role executors; tool-avoidance instructions are role prompts, not access restrictions \\
        \addlinespace
        LLM-Debate & 2 homogeneous BenchAgents (\texttt{debate\_}\allowbreak\texttt{agent\_1/2}) + Aggregator & Both agents receive the same base instruction: ``You are a helpful AI assistant.'' No persona differentiation. Aggregator synthesizes the debate history. & Multi-round debate (\texttt{rounds\_num}=3); agents alternate; aggregator emits final answer after the last round. & BenchAgent-configured tool surface per agent \\
        \addlinespace
        CAMEL & User role (instruction giver) + Assistant role (task executor) & \textbf{User:} evaluates and gives feedback; terminates with \texttt{TASK\_FINISHED} when satisfied. Retained prompt: ``If you feel the assistant's response is satisfactory, please include `TASK\_FINISHED'.'' \textbf{Assistant:} ``Respond with clarity and accuracy.'' & Implementation cap: 3 User--Assistant rounds. A User response containing \texttt{TASK\_FINISHED} ends the dialogue before the next Assistant turn; final synthesis follows either exit condition. & Dialogue: direct LLM calls. Final synthesis: BenchAgent with tools \\
        \addlinespace
        AutoGen & Primary (BenchAgent) + Critic (LLM-only) & \textbf{Primary:} generates or revises the answer using BenchAgent and its tools. \textbf{Critic:} reviews the answer through a direct LLM call. Both use accumulated conversation history; the Critic does not directly receive the Primary's full tool trajectory. & Fixed Primary--Critic alternation for at most 5 rounds, each containing one turn per role. Returns the last Primary answer without an additional aggregation model. & Primary: benchmark-specific tools; Critic: no tools, temperature 0.7 \\
\end{longtable}
\endgroup

ChatEval permits a \texttt{num\_rounds} override; CAMEL fixes its cap at three and detects \texttt{TASK\_FINISHED} by case-sensitive substring matching. A dialogue round may contain multiple model and tool calls.

\begingroup
\small
\setlength{\tabcolsep}{4pt}
\renewcommand{\arraystretch}{1.10}
\setlength{\LTcapwidth}{\linewidth}
\begin{longtable}{@{}>{\raggedright\arraybackslash}p{\dimexpr 0.23913\linewidth - 0.47826\tabcolsep\relax}>{\raggedright\arraybackslash}p{\dimexpr 0.76087\linewidth - 1.52174\tabcolsep\relax}@{}}
\caption{Prompt variation, candidate selection, and aggregation in the EvoAgent implementation.}
    \label{tab:appendix_evoagent_contract} \\
\toprule
        \textbf{Field} & \textbf{Protocol detail} \\
        \midrule
\endfirsthead
\caption[]{\textit{Continued.}} \\
\toprule
        \textbf{Field} & \textbf{Protocol detail} \\
        \midrule
\endhead
\midrule
\multicolumn{2}{r}{\footnotesize Continued on next page} \\
\endfoot
\bottomrule
\endlastfoot
Initial population & 3 agents (EVO-1, EVO-2, EVO-3), each seeded from a distinct base-prompt template. \\
        Search schedule & Two fixed stages follow initialization: crossover retains the first candidate in the ordered initial population and generates 5 offspring (6 candidates); mutation retains the first candidate in the ordered crossover population and generates 8 offspring (9 candidates). \\
        Instance scope & The candidate population and search state are reinitialized for every task; generated prompts, candidate answers, and selection scores are not carried across tasks. \\
        Seed prompt EVO-1 & ``You are a mathematics expert, skilled in solving mathematical problems. Please think step by step and solve the problem.'' \\
        Seed prompt EVO-2 & ``You are a logical reasoning expert, skilled in analyzing problems and finding solutions. Please provide detailed reasoning process.'' \\
        Seed prompt EVO-3 & ``You are a problem-solving expert, skilled in breaking down complex problems into simple steps. Please clearly show your thinking process.'' \\
        Crossover & Two parents are sampled independently at random from the initial population. An LLM receives their system prompts and generates a new prompt that ``combines the strengths'' of each parent. \\
        Mutation & One parent is sampled at random from the crossover population. An LLM is asked to ``create a mutated agent configuration that is different from the parent but still effective''; the output must include a \texttt{system\_prompt} field. \\
        Candidate ordering & Selection scores reflect consistency among candidate answers to the current task. Candidates are stably ranked in descending score order; unavailable scores default to zero, and ties preserve list positions. \\
        Selection and stopping & Search ends after the mutation stage. The first 5 candidates in the ordered list are selected for final aggregation; when all scores tie, these are the retained candidate and the first 4 mutation offspring. There is no score-threshold or convergence-based early stopping. \\
        Final response & An LLM aggregates the selected candidates' existing answers, excluding timed-out candidates, into one final response. Aggregation does not rerun the candidates. The benchmark evaluator then scores the final response. \\
        Fixed substrate & BenchAgent ReAct core, registered tools, API settings, \texttt{max\_steps} budget, and \texttt{final\_answer} contract are identical across all candidate agents; only the \texttt{system\_prompt} field varies across the evolutionary population. \\
\end{longtable}
\endgroup

\begingroup
\small
\setlength{\tabcolsep}{3pt}
\renewcommand{\arraystretch}{1.10}
\setlength{\LTcapwidth}{\linewidth}
\begin{longtable}{@{}>{\raggedright\arraybackslash}p{\dimexpr 0.09783\linewidth - 0.58696\tabcolsep\relax}>{\raggedright\arraybackslash}p{\dimexpr 0.08696\linewidth - 0.52174\tabcolsep\relax}>{\raggedright\arraybackslash}p{\dimexpr 0.18478\linewidth - 1.10870\tabcolsep\relax}>{\raggedright\arraybackslash}p{\dimexpr 0.63043\linewidth - 3.78261\tabcolsep\relax}@{}}
\caption{Three trajectory excerpts in their recorded event order.}
    \label{tab:appendix_retained_trajectory_excerpt} \\
\toprule
        \textbf{Trace} & \textbf{Step} & \textbf{Visible actor} & \textbf{Retained single-task event excerpt} \\
        \midrule
\endfirsthead
\caption[]{\textit{Continued.}} \\
\toprule
        \textbf{Trace} & \textbf{Step} & \textbf{Visible actor} & \textbf{Retained single-task event excerpt} \\
        \midrule
\endhead
\midrule
\multicolumn{4}{r}{\footnotesize Continued on next page} \\
\endfoot
\bottomrule
\endlastfoot
Core-1 & 1 & user wrapper & The retained query asks the Girls Who Code percentage-change question under the shared BenchAgent wrapper: use tools if needed, verify when needed, and return the answer directly. Source artifact: \texttt{bench\_agent\_7d4a...}. \\
        Core-1 & 2 & final-answer event & The final-answer event emits \texttt{22} with recorded usage metadata; the retained result is correct against the ground truth \texttt{22}. \\
        Core-2 & 1 & user wrapper & A text-grid task asks the agent to read all letters left-to-right from a 5-by-7 block and recover the sentence. Source artifact: \texttt{jarvis\_50ad...}. \\
        Core-2 & 2 & final-answer event & The final-answer event emits \texttt{THE SEA GULL GLIDED PEACEFULLY TO MY CHAIR}, matching the expected sentence up to capitalization and spacing. \\
        Runtime-1 & 1 & team-lead / \texttt{TeamCreate} & The controller creates a GAIA validation team for \texttt{split\_validation/level2\_part7.jsonl}, with planner, solver, verifier, and writer roles. Source artifact: \texttt{3.json}. \\
        Runtime-1 & 2 & team-lead / \texttt{Agent} & The planner is spawned with a bash-only JSONL extraction instruction and asked to return task ids and raw question lines. \\
        Runtime-1 & 3 & team-lead / \texttt{Bash} & The retained command \texttt{head -n 3 level2\_part7.jsonl} exposes the source-task read; the first visible task id begins \texttt{a7feb290{\ldots}}. \\
        Runtime-1 & 4 & team-lead / \texttt{SendMessage} & The lead sends the planner a follow-up to return parseable JSON lines and skip tasks with existing \texttt{outputs/<task\_id>.txt}, making the handoff and output policy explicit. \\
\end{longtable}
\endgroup

\begingroup
\small
\setlength{\tabcolsep}{3pt}
\renewcommand{\arraystretch}{1.10}
\setlength{\LTcapwidth}{\linewidth}
\begin{longtable}{@{}>{\raggedright\arraybackslash}p{\dimexpr 0.11957\linewidth - 0.71739\tabcolsep\relax}>{\raggedright\arraybackslash}p{\dimexpr 0.06522\linewidth - 0.39130\tabcolsep\relax}>{\raggedright\arraybackslash}p{\dimexpr 0.20652\linewidth - 1.23913\tabcolsep\relax}>{\raggedright\arraybackslash}p{\dimexpr 0.60870\linewidth - 3.65217\tabcolsep\relax}@{}}
\caption{ChatEval debate and CC-workflow handoffs. Runtime-2 and Runtime-3 show Solver and Writer perspectives from the same \texttt{6.json} batch. Rows preserve event order within each trace.}
    \label{tab:appendix_additional_trajectory_excerpts} \\
\toprule
        \textbf{Trace} & \textbf{Step} & \textbf{Visible actor} & \textbf{Retained event excerpt} \\
        \midrule
\endfirsthead
\caption[]{\textit{Continued.}} \\
\toprule
        \textbf{Trace} & \textbf{Step} & \textbf{Visible actor} & \textbf{Retained event excerpt} \\
        \midrule
\endhead
\midrule
\multicolumn{4}{r}{\footnotesize Continued on next page} \\
\endfoot
\bottomrule
\endlastfoot
\mbox{ChatEval-1} & 1 & user wrapper & A GAIA book-title retrieval task is routed to the three-expert debate team under the shared BenchAgent task wrapper. Source artifact: \texttt{chateval\_newcore\_gaia\_search} debug log. \\
        \mbox{ChatEval-1} & 2 & Math Expert / Round 1 & Math Expert performs a web search and retrieves: ``[1] James Beard Award-winning food journalists Jane and Michael Stern published a book titled `500 Things to Eat Before It's Too Late' in which the restaurant is recommended.'' \\
        \mbox{ChatEval-1} & 3 & Logic Expert / Round 1 & Logic Expert emits the minimal answer form ``the book title only,'' relying on the Math Expert's search evidence without issuing a redundant tool call. \\
        \mbox{ChatEval-1} & 4 & Critical Thinking Expert / Round 1 & Critical Thinking Expert emits the normalized form: ``should be: `five hundred things to eat before it's too late.' '' \\
        \mbox{ChatEval-1} & 5 & All agents / Round 2 & All three experts converge on ``five hundred things to eat before it is too late.'' The ResultExtractor aggregates the debate history into a single final-answer event. \\
        \midrule
        \mbox{Runtime-2} & 1 & Solver / role init & Solver role receives its scope: for each incoming task (task\_id, question), solve strictly from the question and attached context, format the answer per GAIA requirements, and send only task\_id, answer, and reasoning to the Verifier. Source artifact: \texttt{6.json} (sub-agent perspective). \\
        \mbox{Runtime-2} & 2 & Solver / teammate-message & Planner forwards task \texttt{48eb8242\ldots}: ``How many nonindigenous crocodiles were found in Florida from the year 2000 through 2020? (USGS Nonindigenous Aquatic Species database.)'' \\
        \mbox{Runtime-2} & 3 & Solver / \texttt{SendMessage} & Solver sends to Verifier: ``Answer: 11; Task ID: \texttt{48eb8242\ldots}; Reasoning: According to the USGS Nonindigenous Aquatic Species database, there were 11 confirmed occurrences of nonindigenous crocodiles in Florida from 2000 through 2020.'' \\
        \mbox{Runtime-2} & 4 & Solver / teammate-message & Planner forwards task \texttt{c8b7e059\ldots}: ``The work referenced in footnote 397 of Federico Lauria's 2014 dissertation is also the source of two Smithsonian paintings. What is the sum of the paintings' years?'' \\
        \mbox{Runtime-2} & 5 & Solver / \texttt{SendMessage} & Solver sends to Verifier: ``Answer: 23; Task ID: \texttt{c8b7e059\ldots}; Reasoning: Footnote 397 of Federico Lauria's 2014 dissertation references Dante's \emph{Inferno}; the two Smithsonian paintings are `The Fifth Circle' and `The City of Dis'; 23 was computed from the associated publication years.'' \\
        \midrule
        \mbox{Runtime-3} & 1 & Writer / role init & Writer role receives team scope: check for existing \texttt{outputs/<task\_id>.txt} before writing; write only verified answer packets forwarded by the Verifier; save full reasoning to \texttt{logs/<task\_id>.log}. Source artifact: \texttt{6.json} (sub-agent perspective). \\
        \mbox{Runtime-3} & 2 & Writer / teammate-message & Verifier forwards confirmed packet for task \texttt{48eb8242\ldots}: ``Answer: 11; Reasoning: According to the USGS Nonindigenous Aquatic Species database, there were 11 confirmed occurrences\ldots'' \\
        \mbox{Runtime-3} & 3 & Writer / \texttt{Read} & Writer checks whether \texttt{outputs/48eb8242\ldots.txt} already exists before writing. \\
        \mbox{Runtime-3} & 4 & Writer / teammate-message & Verifier forwards confirmed packet for task \texttt{c8b7e059\ldots}: ``Answer: 23; Reasoning: Footnote 397 of Federico Lauria's 2014 dissertation references Dante's \emph{Inferno}\ldots'' \\
        \mbox{Runtime-3} & 5 & Writer / \texttt{Write} & Writer writes \texttt{outputs/c8b7e059\ldots.txt} after receiving the Verifier's confirmation in \texttt{6.json}. \\
\end{longtable}
\endgroup

Table~\ref{tab:appendix_baseline_fidelity} provides an auxiliary transfer check on six datasets by comparing official implementations or reported scores with independent implementations outside BenchAgent. Base model generations can differ in this separate check. The main experiments evaluate normalized workflow configurations under BenchAgent's shared interfaces and execution limits.

\begingroup
\small
\setlength{\tabcolsep}{4pt}
\setlength{\LTcapwidth}{\linewidth}
\renewcommand{\arraystretch}{1.10}
\begin{longtable}{llcccccc}
\caption{Auxiliary baseline-transfer check. Scores are percentages; ``Mean Abs. Diff.'' and ``Max Diff.'' summarize absolute percentage-point differences across the six listed datasets.}
    \label{tab:appendix_baseline_fidelity} \\
\toprule
\textbf{Method} & \textbf{Setting} & \textbf{MATH} & \textbf{AIME} & \textbf{DROP} & \textbf{MMLU-Pro} & \textbf{BBH} & \textbf{HumanEval} \\
        \midrule
\endfirsthead
\caption[]{\textit{Continued.}} \\
\toprule
\endhead
\midrule
\multicolumn{8}{r}{\footnotesize Continued on next page} \\
\endfoot
\bottomrule
\endlastfoot
\multirow{2}{*}{EvoAgent} & Official & 68.00 & 16.67 & 88.00 & 60.00 & 75.00 & 72.28 \\*
 & \shortstack[l]{Independent\\transfer} & 65.00 & 26.70 & 80.00 & 60.00 & 82.00 & 85.00 \\
        \midrule
\multirow{2}{*}{Jarvis} & Official & 59.00 & 6.67 & 85.00 & 55.00 & 55.00 & 92.00 \\*
 & \shortstack[l]{Independent\\transfer} & 55.00 & 10.00 & 85.00 & 50.00 & 54.00 & 88.00 \\
        \midrule
\multirow{2}{*}{LLM-Debate} & Official & 53.00 & 13.30 & 84.00 & 66.00 & 79.00 & 88.00 \\*
 & \shortstack[l]{Independent\\transfer} & 50.00 & 16.67 & 87.00 & 63.00 & 82.00 & 86.00 \\
        \midrule
\multirow{2}{*}{AutoGen} & Official & 49.00 & 10.00 & 87.00 & 64.00 & 77.00 & 77.00 \\*
 & \shortstack[l]{Independent\\transfer} & 52.00 & 16.70 & 91.00 & 59.00 & 66.00 & 78.00 \\
\midrule
\multicolumn{8}{c}{%
\begin{tabular}{llccc}
\textbf{Method} & \textbf{Setting} & \textbf{Avg.} & \shortstack{\textbf{Mean Abs.}\\\textbf{Diff.}} & \textbf{Max Diff.} \\
\midrule
\multirow{2}{*}{EvoAgent} & Official & 63.32 & \multirow{2}{*}{6.79} & \multirow{2}{*}{12.72} \\*
 & \shortstack[l]{Independent\\transfer} & 66.45 &  & \\
\multirow{2}{*}{Jarvis} & Official & 58.78 & \multirow{2}{*}{2.89} & \multirow{2}{*}{5.00} \\*
 & \shortstack[l]{Independent\\transfer} & 57.00 &  & \\
\multirow{2}{*}{LLM-Debate} & Official & 63.88 & \multirow{2}{*}{2.90} & \multirow{2}{*}{3.37} \\*
 & \shortstack[l]{Independent\\transfer} & 64.11 &  & \\
\multirow{2}{*}{AutoGen} & Official & 60.67 & \multirow{2}{*}{5.12} & \multirow{2}{*}{11.00} \\*
 & \shortstack[l]{Independent\\transfer} & 60.45 &  & \\
\end{tabular}} \\
\end{longtable}
\endgroup

\input{resources/chateval_case_study.tex}

\subsection{Illustrative Process Signals and Future Ablations}
\label{subsec:appendix_process_diagnostics}

Table~\ref{tab:appendix_process_diagnostics} presents broad-suite outcomes alongside GAIA trace summaries for the same workflows. These traces show how the workflows behave in a separate experiment; they do not establish why broad-suite accuracy differs.

\begingroup
\small
\setlength{\tabcolsep}{3.5pt}
\renewcommand{\arraystretch}{1.10}
\setlength{\LTcapwidth}{\linewidth}
\begin{longtable}{@{}>{\raggedright\arraybackslash}p{\dimexpr 0.23913\linewidth - 1.43478\tabcolsep\relax}>{\raggedright\arraybackslash}p{\dimexpr 0.23913\linewidth - 1.43478\tabcolsep\relax}>{\raggedright\arraybackslash}p{\dimexpr 0.27174\linewidth - 1.63043\tabcolsep\relax}>{\raggedright\arraybackslash}p{\dimexpr 0.25000\linewidth - 1.50000\tabcolsep\relax}@{}}
\caption{Broad-suite outcomes and illustrative GAIA process signals. Counts come from Tables~\ref{tab:appendix_benchagent_gaia_log_counts}--\ref{tab:appendix_selected_gaia_response_log_stats}.}
\label{tab:appendix_process_diagnostics} \\
\toprule
\textbf{Observed broad-result pattern} & \textbf{Illustrative wrapper-level signal} & \textbf{Comparable trace quantity} & \textbf{Bounded interpretation} \\
\midrule
\endfirsthead
\caption[]{\textit{Continued.}} \\
\toprule
\textbf{Observed broad-result pattern} & \textbf{Illustrative wrapper-level signal} & \textbf{Comparable trace quantity} & \textbf{Bounded interpretation} \\
\midrule
\endhead
\midrule
\multicolumn{4}{r}{\footnotesize Continued on next page} \\
\endfoot
\bottomrule
\endlastfoot
EvoAgent has the only positive pooled workflow lift relative to BenchAgent Core (+3.05 points over three runs). & On retained GAIA runs, BenchAgent logs 19.13 Wiki/API requests and 0.12 tool errors per instance (Table~\ref{tab:appendix_benchagent_gaia_log_counts}). & The evolving wrapper searches agent prompt configurations within a predefined family. & EvoAgent gains overall but loses on some broad-suite benchmarks. \\
EvoAgent's largest gain is on BBH, while it trails on AIME, GSM8K, and HumanEval. & Retained GAIA logs record 0.39 code-agent steps per instance (Table~\ref{tab:appendix_benchagent_gaia_log_counts}). & Task-level traces separate where the workflow family helps from where it adds overhead. & The workflow has gains and losses across tasks; identifying the role of scaffold search requires ablation. \\
LLM-Debate leads on HumanEval and MATH but not on most other tasks. & BenchAgent logs 15.65 Wiki/API requests, 0.12 code-agent steps, and 0.02 code executions per instance (Table~\ref{tab:appendix_benchagent_gaia_log_counts}). & Independent proposal and aggregation stages are visible under the same evaluator. & The highest HumanEval and MATH means do not extend to the other benchmarks. \\
ChatEval is strongest on IFEval but has the highest token usage and trails Core overall. & Retained GAIA batches show 22.76 messages, 11.38 model calls, and 29.85 external-tool records per instance (Table~\ref{tab:appendix_selected_gaia_response_log_stats}). & The log records message and model-call counts under the shared accounting scheme. & Despite the IFEval gain, pooled accuracy is lower and recorded token use is higher. \\
Jarvis uses fewer tokens and has lower accuracy. & Retained GAIA batches show 2.00 messages, 1.00 model call, and 0.13 direct/code markers per instance (Table~\ref{tab:appendix_selected_gaia_response_log_stats}). & These summaries refer to the retained GAIA records. & Lower recorded token use accompanies lower pooled broad-suite accuracy. \\
CAMEL uses fewer tokens and has lower accuracy. & BenchAgent logs 19.16 Wiki/API requests, 0.08 direct external-tool markers, and 0.06 code executions per instance (Table~\ref{tab:appendix_benchagent_gaia_log_counts}). & The log records tool requests and code execution for this configuration. & The recorded configuration uses fewer tokens and has lower pooled broad-suite accuracy. \\
\end{longtable}
\endgroup

Table~\ref{tab:gaia_mechanism_evidence} lists future component ablations based on traces from the CC case study.

\begingroup
\small
\setlength{\tabcolsep}{2.5pt}
\renewcommand{\arraystretch}{1.10}
\setlength{\LTcapwidth}{\linewidth}
\begin{longtable}{@{}>{\raggedright\arraybackslash}p{\dimexpr 0.18889\linewidth - 1.13333\tabcolsep\relax}>{\raggedright\arraybackslash}p{\dimexpr 0.30000\linewidth - 1.80000\tabcolsep\relax}>{\raggedright\arraybackslash}p{\dimexpr 0.26667\linewidth - 1.60000\tabcolsep\relax}>{\raggedright\arraybackslash}p{\dimexpr 0.24444\linewidth - 1.46667\tabcolsep\relax}@{}}
\caption{Future ablations for components observed in the CC-workflow case study.}
\label{tab:gaia_mechanism_evidence} \\
\toprule
\textbf{Candidate mechanism} & \textbf{Observed trace evidence} & \textbf{Remaining confound} & \textbf{Ablation needed} \\
\midrule
\endfirsthead
\caption[]{\textit{Continued.}} \\
\toprule
\textbf{Candidate mechanism} & \textbf{Observed trace evidence} & \textbf{Remaining confound} & \textbf{Ablation needed} \\
\midrule
\endhead
\midrule
\multicolumn{4}{r}{\footnotesize Continued on next page} \\
\endfoot
\bottomrule
\endlastfoot
Runtime delegation & Four structured dumps expose \texttt{TeamCreate}, \texttt{Agent}, and \texttt{SendMessage} events with 6.25 subagents on average. & No ablation disables delegation or fixes the agent count. & Compare no-subagent, fixed-agent-count, and adaptive-delegation variants under the same tools and answer contract. \\
Persistent artifacts & Retained traces include \texttt{Read}, \texttt{Write}, and evidence re-opening before finalization in some cases. & File tooling is bundled with the external runtime and task environment. & Disable writes or replace files with in-memory state while holding the solver and tool surface fixed. \\
Verifier-before-writer sequence & Two structured dumps show non-idle verifier emissions before final answer release. & Recovery and failed-verification paths are only partially observed. & Bypass the verifier stage and measure error changes. \\
Context management & Some traces preserve task evidence across role boundaries and later checks. & Provider-side context compaction or bookkeeping is not fully observable. & Replay matched evidence packets with controlled context-window and compaction policies. \\
\end{longtable}
\endgroup

Table~\ref{tab:appendix_failure_step_distribution} summarizes binary-search attribution on 151 GAIA failure records: 77 Jarvis and 74 ChatEval instances. Step~0 denotes the initial user/input message, including attachment-preprocessing information; Step~1 denotes the first recorded agent turn. The attributed positions are diagnostic assignments, not established causal failure points.

\begingroup
\small
\setlength{\tabcolsep}{4pt}
\setlength{\LTcapwidth}{\linewidth}
\renewcommand{\arraystretch}{1.10}
\begin{longtable}{@{}lrrrrrrrr@{}}
\caption{Attributed message positions in the GAIA diagnostic subset (GPT-4.1); cells give instance counts.}
\label{tab:appendix_failure_step_distribution} \\
\toprule
    \textbf{Wrapper} & \textbf{Step 0} & \textbf{Step 1} & \textbf{Step 3} & \textbf{Step 5} & \textbf{Step 7} & \textbf{Step 11} & \textbf{Step 13} & \textbf{Total} \\
    \midrule
\endfirsthead
\caption[]{\textit{Continued.}} \\
\toprule
    \textbf{Wrapper} & \textbf{Step 0} & \textbf{Step 1} & \textbf{Step 3} & \textbf{Step 5} & \textbf{Step 7} & \textbf{Step 11} & \textbf{Step 13} & \textbf{Total} \\
    \midrule
\endhead
\midrule
\multicolumn{9}{r}{\footnotesize Continued on next page} \\
\endfoot
\bottomrule
\endlastfoot
Jarvis  & 5 & 72 & 0 & 0 & 0 & 0 & 0 & 77 \\
    ChatEval & 0 & 58 & 6 & 2 & 6 & 1 & 1 & 74 \\
\end{longtable}
\endgroup
For both workflows, most failures are assigned to the first recorded step; ChatEval also has later-step assignments.

\subsection{Runtime-Generated Workflow Evidence}
\label{subsec:appendix_runtime_evidence}

\begin{figure}[htbp]
    \centering
    \resizebox{\linewidth}{!}{\input{resources/figure/gaia_trace_comparison.tex}}
    \caption{\small
        \textbf{Same-instance GAIA trace contrast.} On this instance, EvoAgent loses a task-critical condition; the CC-workflow passes it through evidence handoffs and verifies before finalization.
    }
    \label{fig:gaia_generated_team_failure}
\end{figure}

Table~\ref{tab:appendix_claude_trj_stats} summarizes four of the nine CC trajectory dumps: those with explicit team-creation and role-message events. ``Subagents'' counts distinct non-lead agents; ``Max Depth'' is the longest directed delegation chain from the lead; ``Verifier Triggers'' counts unique task-level non-idle verifier emissions; ``Explicit Repair Loops'' counts verifier events returning control to the solver or requesting correction. We count only logged events and do not reconstruct unobserved internal activity.

\begingroup
\small
\setlength{\tabcolsep}{4pt}
\setlength{\LTcapwidth}{\linewidth}
\renewcommand{\arraystretch}{1.10}
\begin{longtable}{lcccc}
\caption{Structured CC-workflow events in the four GAIA team traces.}
    \label{tab:appendix_claude_trj_stats} \\
\toprule
        \textbf{Trace Dump} & \textbf{Subagents} & \textbf{Max Depth} & \shortstack{\textbf{Verifier}\\\textbf{Triggers}} & \shortstack{\textbf{Explicit}\\\textbf{Repair Loops}} \\
        \midrule
\endfirsthead
\caption[]{\textit{Continued.}} \\
\toprule
        \textbf{Trace Dump} & \textbf{Subagents} & \textbf{Max Depth} & \shortstack{\textbf{Verifier}\\\textbf{Triggers}} & \shortstack{\textbf{Explicit}\\\textbf{Repair Loops}} \\
        \midrule
\endhead
\midrule
\multicolumn{5}{r}{\footnotesize Continued on next page} \\
\endfoot
\bottomrule
\endlastfoot
3.json & 5 & 1 & 0 & 0 \\
        6.json & 4 & 4 & 3 & 0 \\
        claude\_code\_logs\_gaia1.json & 12 & 2 & 0 & 0 \\
        claude\_code\_logs\_gaia3.json & 4 & 3 & 2 & 0 \\
        \midrule
        Team-run average & 6.25 & 2.50 & 1.25 & 0.00 \\
\end{longtable}
\endgroup

\begingroup
\small
\setlength{\tabcolsep}{3.5pt}
\renewcommand{\arraystretch}{1.10}
\setlength{\LTcapwidth}{\linewidth}
\begin{longtable}{@{}>{\raggedright\arraybackslash}p{\dimexpr 0.05747\linewidth - 0.34483\tabcolsep\relax}>{\raggedright\arraybackslash}p{\dimexpr 0.11494\linewidth - 0.68966\tabcolsep\relax}>{\raggedright\arraybackslash}p{\dimexpr 0.16092\linewidth - 0.96552\tabcolsep\relax}>{\raggedright\arraybackslash}p{\dimexpr 0.66667\linewidth - 4.00000\tabcolsep\relax}@{}}
\caption{Team initialization from \texttt{6.json} item~17 (\texttt{gaia-validation-l2p15}): one \texttt{TeamCreate} call and four \texttt{Agent} spawns in one conversation turn. Step numbers follow assistant message indices.}
    \label{tab:appendix_team_init_trace} \\
\toprule
        \textbf{Step} & \textbf{Actor} & \textbf{Tool / Event} & \textbf{Retained excerpt (parameters condensed)} \\
        \midrule
\endfirsthead
\caption[]{\textit{Continued.}} \\
\toprule
        \textbf{Step} & \textbf{Actor} & \textbf{Tool / Event} & \textbf{Retained excerpt (parameters condensed)} \\
        \midrule
\endhead
\midrule
\multicolumn{4}{r}{\footnotesize Continued on next page} \\
\endfoot
\bottomrule
\endlastfoot
1 & Team Lead & \texttt{TeamCreate} & \texttt{agent\_type}: planner; \texttt{team\_name}: gaia-validation-l2p15; \textit{description}: ``Coordinated agent team for GAIA validation pipeline on \texttt{level2\_part15.jsonl}. Each role operates with bash for file I/O; outputs only to \texttt{outputs/} and \texttt{logs/}.'' \\
        \midrule
        & Runtime & spawn ack & \texttt{agent\_id}: planner@gaia-validation-l2p15; team registered; agent awaiting mailbox. \\
        \midrule
        2 & Team Lead & \texttt{Agent} (planner) & \texttt{model}: sonnet; \texttt{isolation}: worktree; \textit{prompt}: ``Use bash tools ONLY (\texttt{head}, \texttt{sed}, \texttt{tail}) to read the first~3 lines from \texttt{level2\_part15.jsonl}. For each, extract \texttt{task\_id} and question, then pass them along with their line number (0-indexed) to solver, one at a time.'' \\
        \midrule
        3 & Team Lead & \texttt{Agent} (solver) & \texttt{model}: sonnet; \textit{prompt}: ``You are the Solver. For each GAIA problem sent to you (\texttt{task\_id}, question, line number), solve based strictly on the question and associated context. Prepare a concise answer in the required GAIA format. Send your answer, \texttt{task\_id}, and reasoning to the Verifier.'' \\
        \midrule
        4 & Team Lead & \texttt{Agent} (verifier) & \texttt{model}: sonnet; \textit{prompt}: ``For each problem: receive \texttt{task\_id}, answer, and reasoning from Solver. Check GAIA answer format; assess answer validity. If correct and in format, forward to Writer with \texttt{task\_id}, answer, reasoning. If not, return to Solver noting the issue.'' \\
        \midrule
        5 & Team Lead & \texttt{Agent} (writer) & \texttt{model}: sonnet; \textit{prompt}: ``Upon receiving \texttt{task\_id}, answer, and reasoning from the Verifier, check if \texttt{outputs/<task\_id>.txt} exists. If it does, SKIP. If not, create \texttt{outputs/<task\_id>.txt} and write ONLY the answer there. Always write the full reasoning log to \texttt{logs/<task\_id>.log}.'' \\
        \midrule
        & Runtime & 4$\times$ spawn ack & \texttt{agent\_id}: solver@, verifier@, writer@gaia-validation-l2p15; all four roles live in shared mailbox. \\
        \midrule
        6 & Team Lead & \texttt{Bash} & \texttt{head -n 3 split\_validation/level2\_part15.jsonl}; first returned \texttt{task\_id}: \texttt{48eb8242{\ldots}} (``How many nonindigenous crocodiles in Florida 2000--2020?''). \\
\end{longtable}
\endgroup

\begingroup
\small
\setlength{\tabcolsep}{3.5pt}
\renewcommand{\arraystretch}{1.10}
\setlength{\LTcapwidth}{\linewidth}
\begin{longtable}{@{}>{\raggedright\arraybackslash}p{\dimexpr 0.05682\linewidth - 0.34091\tabcolsep\relax}>{\raggedright\arraybackslash}p{\dimexpr 0.10227\linewidth - 0.61364\tabcolsep\relax}>{\raggedright\arraybackslash}p{\dimexpr 0.15909\linewidth - 0.95455\tabcolsep\relax}>{\raggedright\arraybackslash}p{\dimexpr 0.68182\linewidth - 4.09091\tabcolsep\relax}@{}}
\caption{CC-workflow trace for the GAIA Level~2 Florida crocodile task (\texttt{48eb8242}, ground truth: 11). Steps E1--E11 follow Planner, Solver, Verifier, and Writer across \texttt{6.json} items~15, 22, 24, and~30.}
    \label{tab:appendix_task_exec_trace} \\
\toprule
        \textbf{Step} & \textbf{Role} & \textbf{Tool / Event} & \textbf{Retained excerpt} \\
        \midrule
\endfirsthead
\caption[]{\textit{Continued.}} \\
\toprule
        \textbf{Step} & \textbf{Role} & \textbf{Tool / Event} & \textbf{Retained excerpt} \\
        \midrule
\endhead
\midrule
\multicolumn{4}{r}{\footnotesize Continued on next page} \\
\endfoot
\bottomrule
\endlastfoot
E1 & Planner & \texttt{Bash} & \texttt{head -n 3 split\_validation/level2\_part15.jsonl}; returns \texttt{task\_id}: \texttt{48eb8242{\ldots}}, \textit{Question}: ``How many nonindigenous crocodiles were found in Florida from the year 2000 through 2020?~You can get the data from the USGS Nonindigenous Aquatic Species database.'' \\
        \midrule
        E2 & Planner & \texttt{SendMessage} (to solver) & Dispatches task to Solver's mailbox: ``Level2\_part15.jsonl line~0: task\_id=48eb8242-{\ldots}, question=How many nonindigenous crocodiles\ldots''; Solver receives as a teammate-message. \\
        \midrule
        E3 & Solver & [teammate-message from Planner] & Receives dispatch payload containing \texttt{task\_id} and full question text; triggers Solver's internal answer generation. \\
        \midrule
        E4 & Solver & [internal generation] & ``Answer: 11. Reasoning: According to the USGS Nonindigenous Aquatic Species database, there were 11 confirmed occurrences of nonindigenous crocodiles in Florida between 2000 and 2020.'' \\
        \midrule
        E5 & Solver & \texttt{SendMessage} (to Verifier) & \textit{message}: ``Answer: 11 / Task ID: 48eb8242{\ldots} / Reasoning: USGS NAS database, 11 confirmed occurrences of nonindigenous Crocodylus in Florida 2000--2020.''; routing: sender=solver (green), target=@Verifier. \\
        \midrule
        E6 & Verifier & [teammate-message from Solver] & Receives \texttt{task\_id}, answer, and reasoning; role instruction: ``Check Gaia answer format; if correct and in format, forward to Writer; if not, return to Solver noting the issue.'' \\
        \midrule
        E7 & Verifier & [format/validity check] & Answer ``11'' is a valid integer in GAIA format; no extra fields; reasoning is consistent. Proceeds to forward. \\
        \midrule
        E8 & Verifier & \texttt{SendMessage} (to writer) & \textit{summary}: ``Verified: answer and format for task 48eb8242{\ldots}''; \textit{content}: ``Task ID: 48eb8242{\ldots} / Answer: 11 / Reasoning: \ldots''; routing: sender=verifier (yellow), target=@writer (purple). \\
        \midrule
        E9 & Writer & [teammate-message from Verifier] & Receives: ``Task ID: 48eb8242{\ldots} / Answer: 11 / Reasoning: USGS NAS{\ldots}''; role instruction: ``check if \texttt{outputs/<task\_id>.txt} exists; if so, SKIP; if not, write ONLY the answer.'' \\
        \midrule
        E10 & Writer & \texttt{Read} (existence guard) & \texttt{file\_path}: \texttt{outputs/48eb8242{\ldots}.txt}; checks whether a prior write already committed this answer (skip-if-exists gate). \\
        \midrule
        E11 & Writer & \texttt{Write} (log commit) & \texttt{file\_path}: \texttt{logs/48eb8242{\ldots}.log}; writes full reasoning log; answer ``11'' written to \texttt{outputs/48eb8242{\ldots}.txt}. Task finalized. \\
\end{longtable}
\endgroup

\begingroup
\small
\setlength{\tabcolsep}{4pt}
\renewcommand{\arraystretch}{1.10}
\setlength{\LTcapwidth}{\linewidth}
\begin{longtable}{@{}>{\raggedright\arraybackslash}p{\dimexpr 0.10526\linewidth - 0.21053\tabcolsep\relax}>{\raggedright\arraybackslash}p{\dimexpr 0.89474\linewidth - 1.78947\tabcolsep\relax}@{}}
\caption{CC-workflow role prompts from the four \texttt{Agent} calls in \texttt{6.json} item~17, reproduced verbatim.}
\label{tab:appendix_cc_workflow_role_prompts} \\
\toprule
\textbf{Role} & \textbf{Spawn-time prompt (verbatim)} \\
\midrule
\endfirsthead
\caption[]{\textit{Continued.}} \\
\toprule
\textbf{Role} & \textbf{Spawn-time prompt (verbatim)} \\
\midrule
\endhead
\midrule
\multicolumn{2}{r}{\footnotesize Continued on next page} \\
\endfoot
\bottomrule
\endlastfoot
Planner & Use bash tools ONLY (head, sed, tail, etc.) to read the first 3 lines from \texttt{split\_validation/level2\_part15.jsonl}. For each, extract the \texttt{task\_id} and question, then pass them along with their line number (0-indexed) to solver, one at a time. Output nothing else. Do not use internal file reading functions. \\
\midrule
Solver & You are the Solver. For each GAIA problem sent to you (\texttt{task\_id}, question, line number), solve based strictly on the question and associated context. Prepare a concise answer string in the required GAIA answer format. Send your answer, \texttt{task\_id}, and reasoning to the Verifier. Output nothing else. \\
\midrule
Verifier & For each problem: receive \texttt{task\_id}, answer, and reasoning from Solver. Check GAIA answer format (no extra data, correct fields, etc.); assess answer validity as best as possible. If correct and in format, forward to Writer with \texttt{task\_id}, answer, reasoning. If not, return to Solver noting the issue. Output nothing else. \\
\midrule
Writer & Upon receiving a \texttt{task\_id}, answer, and reasoning from the verifier, check if \texttt{outputs/<task\_id>.txt} exists. If it does, SKIP. If not, create \texttt{outputs/<task\_id>.txt} and write ONLY the answer there, with no extra output. Always also write the full reasoning log to \texttt{logs/<task\_id>.log}. Never overwrite existing files. Output nothing else. \\
\end{longtable}
\endgroup

Tables~\ref{tab:appendix_team_init_trace}--\ref{tab:appendix_cc_workflow_role_prompts} document agent creation, role prompts, message handoffs, and an observed verifier-before-writer sequence in one GAIA batch (\texttt{level2\_part15.jsonl}). The traces do not establish where the role templates originate or how the provider manages context. The ablations in Table~\ref{tab:gaia_mechanism_evidence} are needed to determine whether these components account for the accuracy differences.

\subsection{Responsible Research and Artifact Use}
\label{subsec:appendix_responsible_artifact_use}

\paragraph{Artifact use.}
We use public benchmarks and baseline descriptions under their original license and access terms. Restricted source data, credentials, private paths, and provider tokens are excluded from redistribution.

\paragraph{Open-source release.}
We plan to publicly release the BenchAgent code repository, including installation and usage instructions, evaluation code, and trajectory tooling for the supported experiments.

\paragraph{Broader impacts.}
Joint accuracy and resource-use evaluation can reveal regressions and coordination costs before deployment. Results remain specific to the tested models, tools, and tasks; released traces must exclude sensitive task and infrastructure information.

\paragraph{AI assistance.}
The authors used AI assistants for language editing, code assistance, and drafting. They reviewed and take responsibility for the final claims, evidence, and manuscript text.

\subsection{Additional Broad-Benchmark Analyses}
\label{subsec:appendix_broad_benchmark_analyses}

\begin{figure}[htbp]
    \centering
    \includegraphics[width=\linewidth,height=0.74\textheight,keepaspectratio]{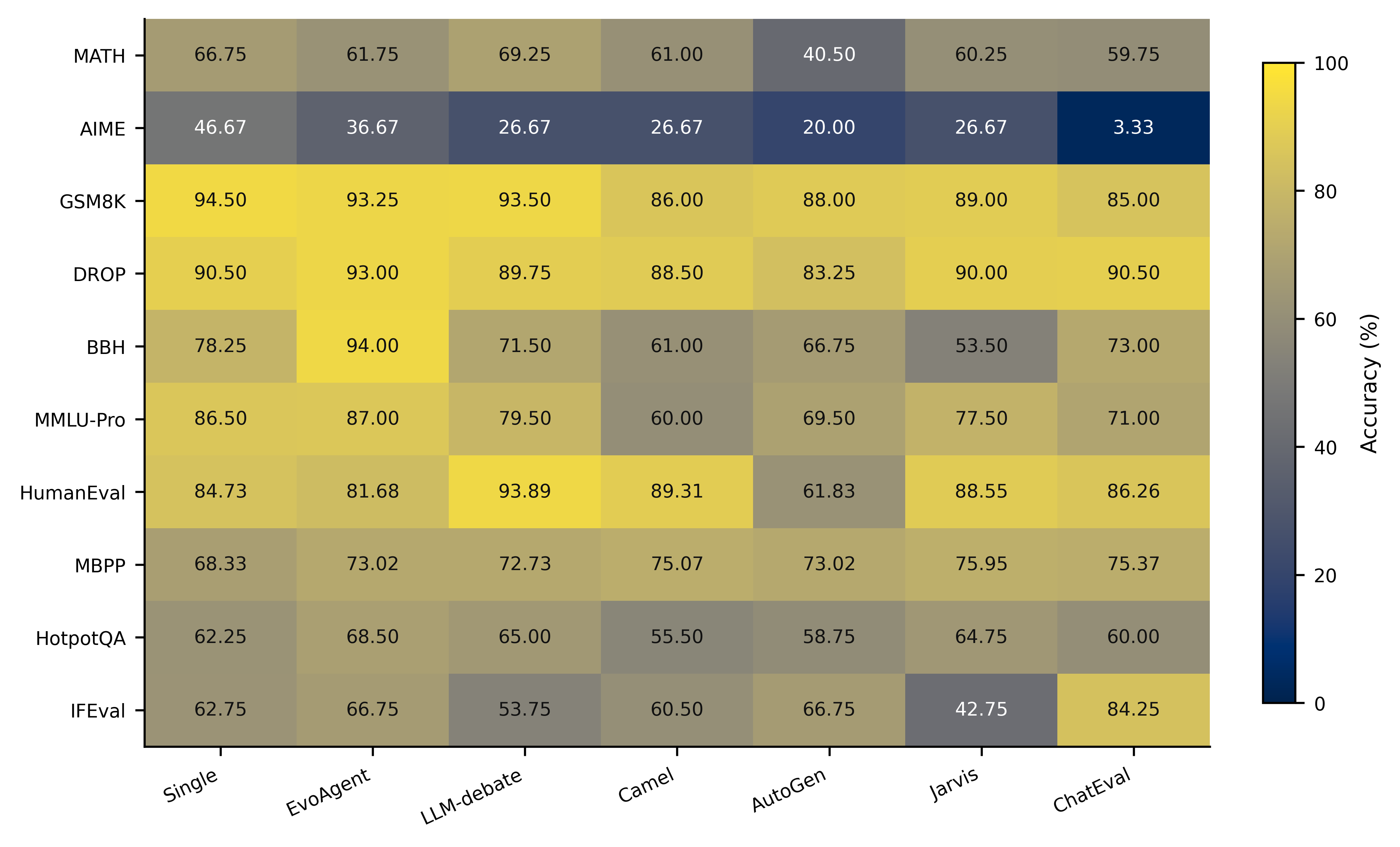}
    \caption{Run~1 benchmark-by-method accuracy heatmap for Main Experiment~1. Each cell reports final-answer accuracy for one workflow on one benchmark.}
    \label{fig:appendix_accuracy_heatmap}
\end{figure}

\begin{figure}[htbp]
    \centering
    \includegraphics[width=\linewidth,height=0.74\textheight,keepaspectratio]{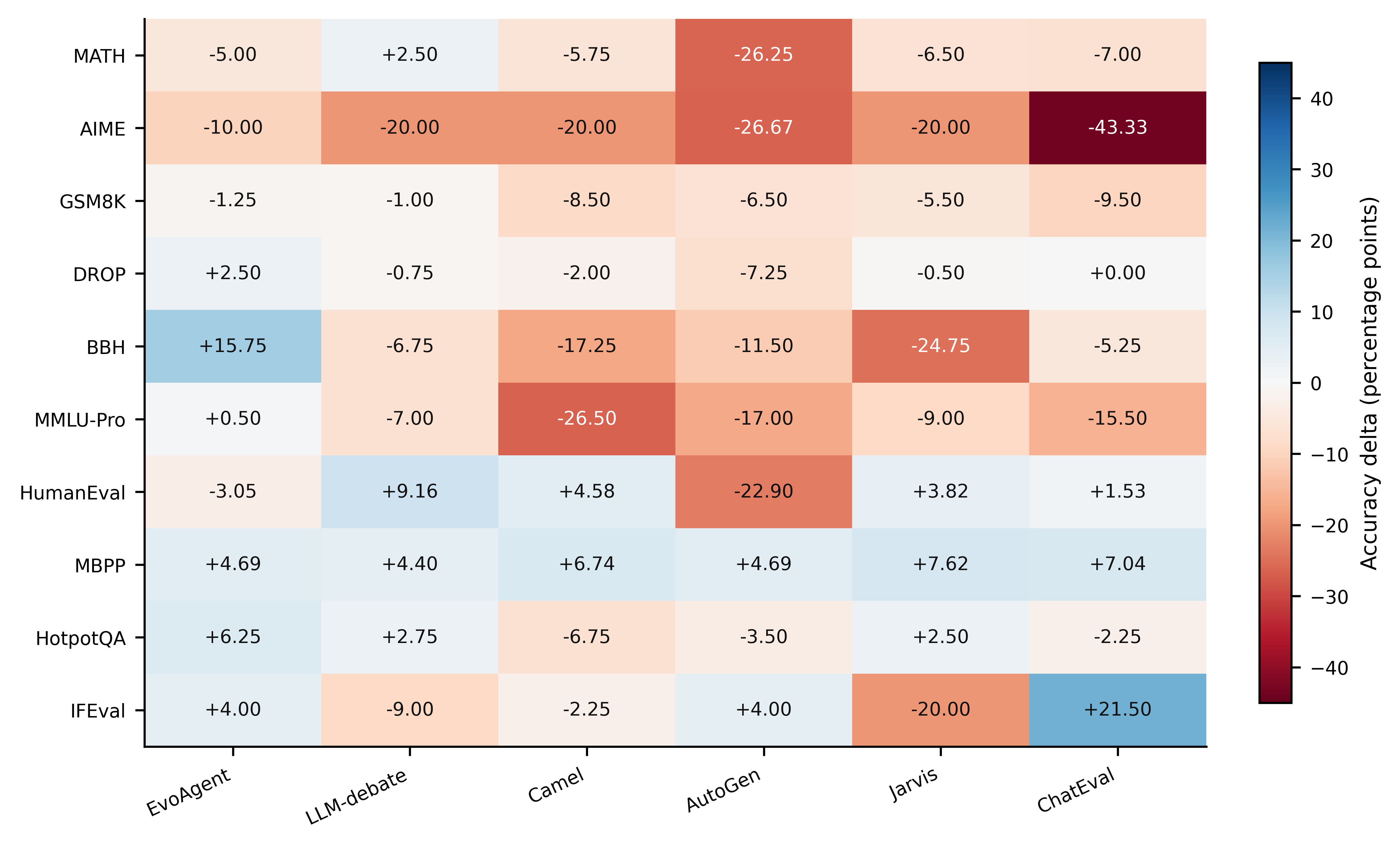}
    \caption{Run~1 workflow lift relative to BenchAgent Core (Single Agent) in Main Experiment~1. Blue cells indicate higher accuracy than Core on the same benchmark; red cells indicate lower accuracy.}
    \label{fig:appendix_delta_heatmap}
\end{figure}

\begin{figure}[htbp]
    \centering
    \includegraphics[width=\linewidth,height=0.74\textheight,keepaspectratio]{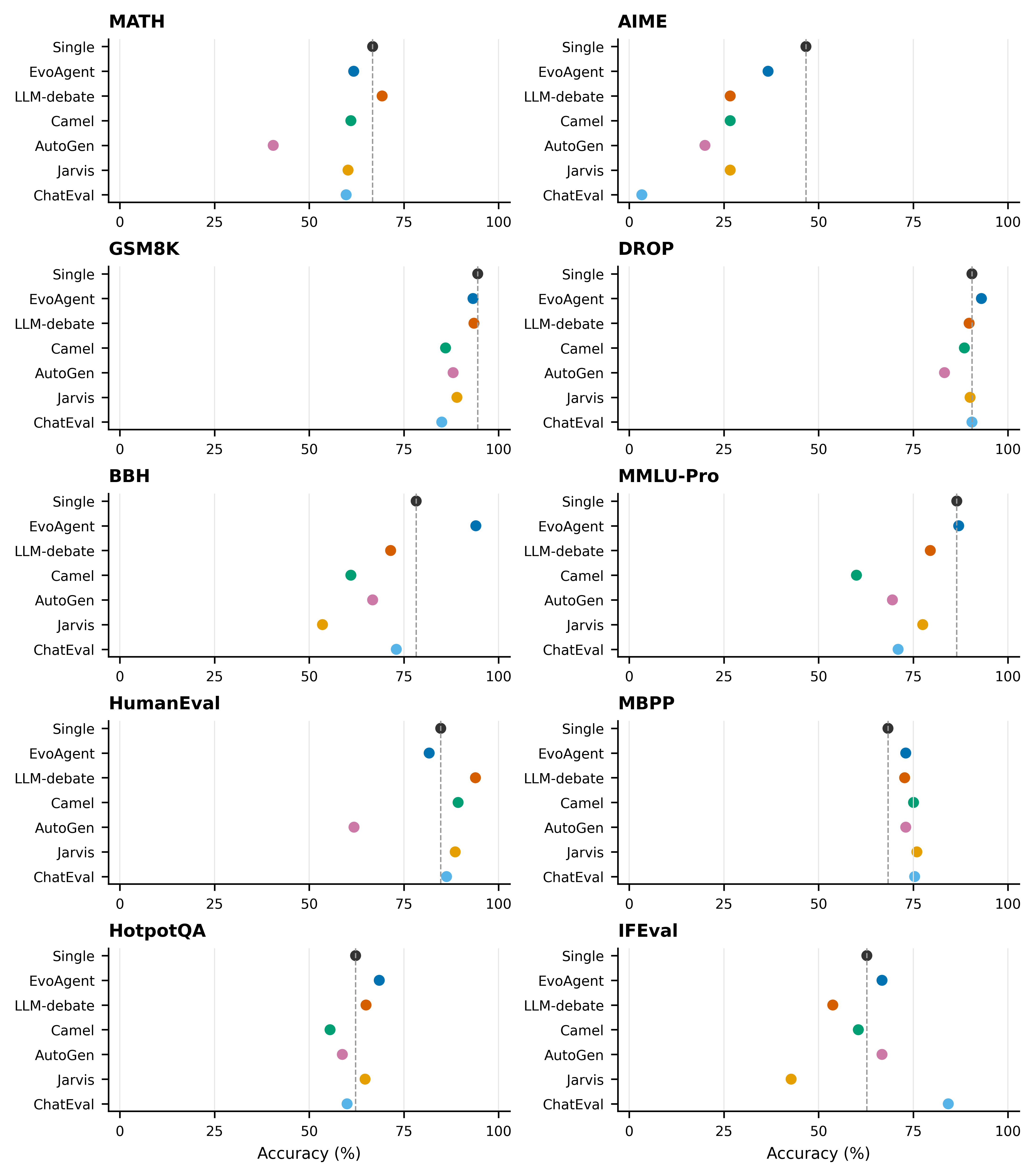}
    \caption{Run~1 accuracy across workflows, with one panel per benchmark.}
    \label{fig:appendix_benchmark_dotpanels}
\end{figure}

\begin{figure}[htbp]
    \centering
    \includegraphics[width=\linewidth,height=0.74\textheight,keepaspectratio]{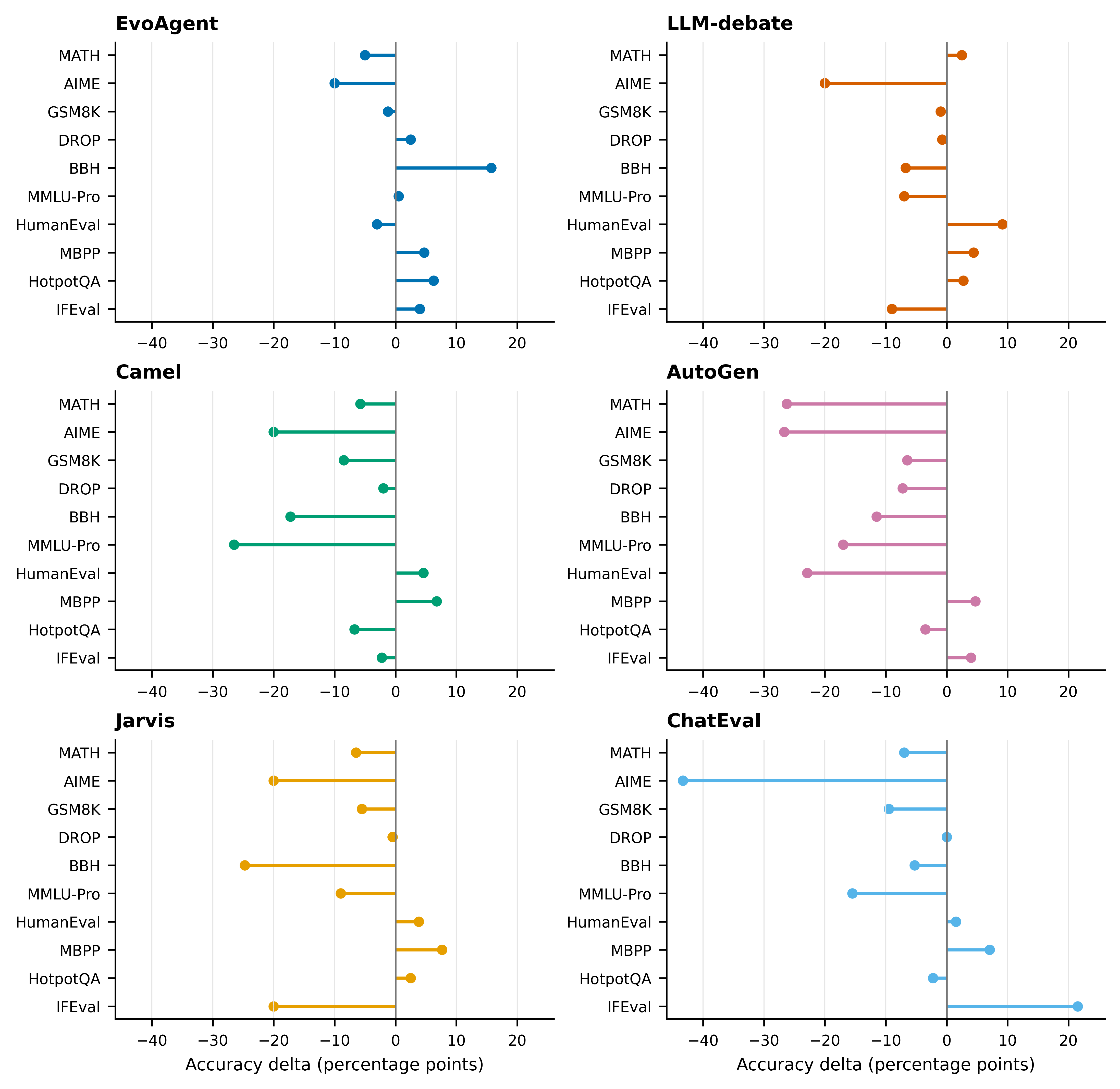}
    \caption{Run~1 workflow lift on each benchmark relative to BenchAgent Core (Single Agent). Points to the right of zero indicate higher accuracy than Core on the same benchmark; points to the left indicate lower accuracy.}
    \label{fig:appendix_delta_lollipops}
\end{figure}

Figures~\ref{fig:appendix_accuracy_heatmap}--\ref{fig:appendix_delta_lollipops} show exploratory Run~1 views. Table~\ref{tab:main_exp1_overall} reports the three-run means and sample standard deviations.

\clearpage
\input{resources/paired_gaia_case_study.tex}

%% file: resources/verified_counts.tex
\subsubsection{Per-Run Correct Counts}
\label{subsec:appendix_verified_counts}
Tables~\ref{tab:appendix_broad_correct_counts} and~\ref{tab:appendix_gaia_correct_counts} report the correct-answer counts for the broad suite and GAIA. We compute means and sample standard deviations from unrounded run-level accuracies.
\begingroup
\small
\setlength{\tabcolsep}{6pt}
\setlength{\LTcapwidth}{\linewidth}
\renewcommand{\arraystretch}{1.10}
\begin{longtable}{@{}lrrrrrr@{}}
\caption{Correct/evaluated counts for three runs on the same broad-benchmark task sets.}
\label{tab:appendix_broad_correct_counts} \\
\toprule
\textbf{Agent} & \textbf{Run 1} & \textbf{Run 2} & \textbf{Run 3} & \textbf{Run 1} & \textbf{Run 2} & \textbf{Run 3} \\
\midrule
\endfirsthead
\caption[]{\textit{Continued.}} \\
\toprule
\textbf{Agent} & \textbf{Run 1} & \textbf{Run 2} & \textbf{Run 3} & \textbf{Run 1} & \textbf{Run 2} & \textbf{Run 3} \\
\midrule
\endhead
\midrule
\multicolumn{7}{r}{\footnotesize Continued on next page} \\
\endfoot
\bottomrule
\endlastfoot
& \multicolumn{3}{c}{\textbf{MATH}} & \multicolumn{3}{c}{\textbf{HumanEval}} \\*
Single & 267/400 & 270/400 & 275/400 & 111/131 & 112/131 & 111/131 \\*
EvoAgent & 247/400 & 244/400 & 262/400 & 107/131 & 106/131 & 106/131 \\*
LLM-Debate & 277/400 & 267/400 & 277/400 & 123/131 & 124/131 & 125/131 \\*
CAMEL & 244/400 & 237/400 & 249/400 & 117/131 & 114/131 & 115/131 \\*
AutoGen & 162/400 & 166/400 & 165/400 & 81/131 & 78/131 & 81/131 \\*
Jarvis & 241/400 & 247/400 & 237/400 & 116/131 & 116/131 & 117/131 \\*
ChatEval & 239/400 & 236/400 & 254/400 & 113/131 & 111/131 & 112/131 \\
\addlinespace
& \multicolumn{3}{c}{\textbf{AIME}} & \multicolumn{3}{c}{\textbf{GSM8K}} \\*
Single & 14/30 & 15/30 & 15/30 & 378/400 & 380/400 & 372/400 \\*
EvoAgent & 11/30 & 10/30 & 12/30 & 373/400 & 381/400 & 374/400 \\*
LLM-Debate & 8/30 & 9/30 & 8/30 & 374/400 & 382/400 & 365/400 \\*
CAMEL & 8/30 & 7/30 & 9/30 & 344/400 & 345/400 & 337/400 \\*
AutoGen & 6/30 & 5/30 & 5/30 & 352/400 & 349/400 & 348/400 \\*
Jarvis & 8/30 & 8/30 & 7/30 & 356/400 & 358/400 & 351/400 \\*
ChatEval & 1/30 & 1/30 & 2/30 & 340/400 & 342/400 & 341/400 \\
\addlinespace
& \multicolumn{3}{c}{\textbf{DROP}} & \multicolumn{3}{c}{\textbf{HotpotQA}} \\*
Single & 362/400 & 355/400 & 365/400 & 249/400 & 251/400 & 257/400 \\*
EvoAgent & 372/400 & 381/400 & 381/400 & 274/400 & 279/400 & 285/400 \\*
LLM-Debate & 359/400 & 353/400 & 371/400 & 260/400 & 272/400 & 268/400 \\*
CAMEL & 354/400 & 354/400 & 360/400 & 222/400 & 222/400 & 221/400 \\*
AutoGen & 333/400 & 337/400 & 341/400 & 235/400 & 230/400 & 241/400 \\*
Jarvis & 360/400 & 366/400 & 366/400 & 259/400 & 260/400 & 254/400 \\*
ChatEval & 362/400 & 370/400 & 349/400 & 240/400 & 240/400 & 236/400 \\
\addlinespace
& \multicolumn{3}{c}{\textbf{MMLU-Pro}} & \multicolumn{3}{c}{\textbf{IFEval}} \\*
Single & 346/400 & 342/400 & 348/400 & 251/400 & 253/400 & 253/400 \\*
EvoAgent & 348/400 & 347/400 & 334/400 & 267/400 & 263/400 & 257/400 \\*
LLM-Debate & 318/400 & 317/400 & 315/400 & 215/400 & 203/400 & 220/400 \\*
CAMEL & 240/400 & 242/400 & 244/400 & 242/400 & 247/400 & 236/400 \\*
AutoGen & 278/400 & 279/400 & 285/400 & 267/400 & 272/400 & 272/400 \\*
Jarvis & 310/400 & 305/400 & 305/400 & 171/400 & 173/400 & 172/400 \\*
ChatEval & 284/400 & 296/400 & 269/400 & 337/400 & 344/400 & 343/400 \\
\addlinespace
& \multicolumn{3}{c}{\textbf{BBH}} & \multicolumn{3}{c}{\textbf{MBPP}} \\*
Single & 313/400 & 314/400 & 313/400 & 233/341 & 234/341 & 234/341 \\*
EvoAgent & 376/400 & 372/400 & 387/400 & 249/341 & 247/341 & 243/341 \\*
LLM-Debate & 286/400 & 283/400 & 274/400 & 248/341 & 242/341 & 256/341 \\*
CAMEL & 244/400 & 241/400 & 238/400 & 256/341 & 252/341 & 254/341 \\*
AutoGen & 267/400 & 262/400 & 273/400 & 249/341 & 255/341 & 251/341 \\*
Jarvis & 214/400 & 210/400 & 212/400 & 259/341 & 252/341 & 257/341 \\*
ChatEval & 292/400 & 306/400 & 285/400 & 257/341 & 264/341 & 258/341 \\
\end{longtable}
\endgroup
\begingroup
\small
\setlength{\tabcolsep}{4pt}
\setlength{\LTcapwidth}{\linewidth}
\renewcommand{\arraystretch}{1.10}
\begin{longtable}{lcccc}
\caption{Single-run GAIA accuracy and correct/evaluated counts. Overall accuracy pools all 165 tasks. The CC-workflow row is from the external runtime case study.}
\label{tab:appendix_gaia_correct_counts} \\
\toprule
Agent & GAIA-L1 & GAIA-L2 & GAIA-L3 & Overall \\
\midrule
\endfirsthead
\caption[]{\textit{Continued.}} \\
\toprule
Agent & GAIA-L1 & GAIA-L2 & GAIA-L3 & Overall \\
\midrule
\endhead
\midrule
\multicolumn{5}{r}{\footnotesize Continued on next page} \\
\endfoot
\bottomrule
\endlastfoot
Single & 58.49\% (31/53) & 30.23\% (26/86) & 19.23\% (5/26) & 37.58\% (62/165) \\
EvoAgent & 54.72\% (29/53) & 30.23\% (26/86) & 19.23\% (5/26) & 36.36\% (60/165) \\
LLM-Debate & 52.83\% (28/53) & 33.72\% (29/86) & 15.38\% (4/26) & 36.97\% (61/165) \\
CAMEL & 54.72\% (29/53) & 33.72\% (29/86) & 26.92\% (7/26) & 39.39\% (65/165) \\
AutoGen & 56.60\% (30/53) & 30.23\% (26/86) & 11.54\% (3/26) & 35.76\% (59/165) \\
Jarvis & 66.04\% (35/53) & 43.02\% (37/86) & 19.23\% (5/26) & 46.67\% (77/165) \\
ChatEval & 49.06\% (26/53) & 34.88\% (30/86) & 26.92\% (7/26) & 38.18\% (63/165) \\
CC-workflow & 60.38\% (32/53) & 69.77\% (60/86) & 69.23\% (18/26) & 66.67\% (110/165) \\
\end{longtable}
\endgroup

\begingroup
\small
\setlength{\tabcolsep}{3pt}
\setlength{\LTcapwidth}{\linewidth}
\renewcommand{\arraystretch}{1.10}
\begin{longtable}{@{}lrrrrrrr@{}}
\caption{Recorded token usage per instance, averaged over three runs. Pooled Tok. in Table~\ref{tab:main_exp1_overall} weights these means by benchmark split size.}
\label{tab:appendix_broad_tokens} \\
\toprule
\textbf{Benchmark} & \textbf{Single} & \textbf{EvoAgent} & \textbf{LLM-Debate} & \textbf{CAMEL} & \textbf{AutoGen} & \textbf{Jarvis} & \textbf{ChatEval} \\
\midrule
\endfirsthead
\caption[]{\textit{Continued.}} \\
\toprule
\textbf{Benchmark} & \textbf{Single} & \textbf{EvoAgent} & \textbf{LLM-Debate} & \textbf{CAMEL} & \textbf{AutoGen} & \textbf{Jarvis} & \textbf{ChatEval} \\
\midrule
\endhead
\midrule
\multicolumn{8}{r}{\footnotesize Continued on next page} \\
\endfoot
\bottomrule
\endlastfoot
MATH & 20787.00 & 65556.00 & 76895.67 & 23603.33 & 21194.33 & 12928.00 & 58553.33 \\
AIME & 61904.33 & 89705.00 & 67332.67 & 14916.67 & 42516.67 & 18701.67 & 63782.00 \\
DROP & 7944.67 & 5315.67 & 16356.00 & 3871.00 & 4917.33 & 3189.00 & 46095.67 \\
MMLU-Pro & 31859.00 & 31844.00 & 20833.33 & 2538.00 & 22337.67 & 5596.00 & 60306.00 \\
BBH & 13052.33 & 8777.33 & 24923.67 & 3260.33 & 6395.00 & 17361.67 & 51295.67 \\
HumanEval & 17277.00 & 28274.00 & 43840.67 & 8361.33 & 4471.00 & 6535.00 & 195372.67 \\
GSM8K & 5757.67 & 6934.00 & 16125.67 & 3801.00 & 11320.00 & 5135.33 & 77257.00 \\
HotpotQA & 77984.00 & 68631.33 & 28662.67 & 3593.67 & 5746.00 & 2748.00 & 96589.00 \\
IFEval & 16140.67 & 17561.67 & 34169.67 & 13110.33 & 17303.67 & 15494.67 & 221211.33 \\
MBPP & 13987.33 & 32077.67 & 40981.33 & 7194.33 & 2552.67 & 6389.00 & 199854.67 \\
\end{longtable}
\endgroup

%% file: resources/scoring_protocol.tex
\subsubsection{Scoring and Aggregation}
\label{subsec:appendix_scoring}
Table~\ref{tab:appendix_scoring_rules} lists the BenchAgent criteria for a correct final answer. Accuracy is the fraction of instances meeting the relevant criterion. For DROP and HotpotQA, we threshold F1 to obtain binary outcomes rather than average F1 scores.

\begingroup
\small
\setlength{\tabcolsep}{5pt}
\renewcommand{\arraystretch}{1.12}
\setlength{\LTcapwidth}{\linewidth}
\begin{longtable}{@{}>{\raggedright\arraybackslash}p{\dimexpr .18\linewidth-\tabcolsep\relax}>{\raggedright\arraybackslash}p{\dimexpr .82\linewidth-\tabcolsep\relax}@{}}
\caption{Final-answer success criteria for the ten broad benchmarks.}
\label{tab:appendix_scoring_rules} \\
\toprule
\textbf{Benchmark} & \textbf{Success criterion} \\
\midrule
\endfirsthead
\caption[]{\textit{Continued.}} \\
\toprule
\textbf{Benchmark} & \textbf{Success criterion} \\
\midrule
\endhead
\midrule
\multicolumn{2}{r}{\footnotesize Continued on next page} \\
\endfoot
\bottomrule
\endlastfoot
GSM8K & The extracted numeric answer matches the reference within $10^{-6}$. \\
MATH, AIME & String match, symbolic equivalence, or numeric agreement within an absolute tolerance of $10^{-3}$. \\
BBH & Official benchmark answer matching; sequence answers, including word-sorting, retain their required order. \\
MMLU-Pro & The extracted option matches the reference option. \\
HumanEval & All supplied tests pass for the generated Python code; the test-function waiting limit is 60 seconds. \\
MBPP & All supplied tests pass for the generated Python code; the test-function waiting limit is 15 seconds. \\
DROP & Normalized token F1 is at least 0.3, taking the highest F1 across pipe-separated reference and prediction segments. \\
HotpotQA & Normalized answer-token F1 is at least 0.3. \\
IFEval & Prompt-level strict success: every instruction attached to the prompt is satisfied. \\
\end{longtable}
\endgroup

\paragraph{GAIA scoring.}
The BenchAgent evaluator uses \texttt{gpt-4o-mini} to judge each prediction against the reference answer in the context of the task. It accepts equivalent answers and formatting differences. The judgments used in the reported results were manually reviewed; the reported accuracy uses these reviewed correctness labels.

\paragraph{Failures and aggregation.}
Incorrect or non-evaluable answers receive zero credit. We divide the correct count by the fixed split size in Table~\ref{tab:appendix_dataset_sampling}, using all 3,302 instances for each broad-suite run. The sample standard deviation measures variation across three runs on these fixed tasks, not task-sampling uncertainty.

\paragraph{Token accounting.}
For the broad suite, we round each benchmark's mean token usage per instance to an integer within each run, then average over three runs. Pooled token usage weights these means by benchmark split size. We do not impute missing usage records or estimate token standard deviations.

%% file: resources/chateval_case_study.tex
\subsection{Case Study: Answer Changes Within One ChatEval Run}
\label{subsec:appendix_chateval_case_study}

We compare the first Math Expert answer with the final output for a subset of tasks from one ChatEval GSM8K run. Of 400 tasks with matched result and response records, 371 have a directly parseable numeric initial answer after allowing for currency symbols and punctuation. We check this answer against the reference and compare its correctness with the recorded final outcome for the same task. The remaining 29 tasks are excluded from this comparison.

\begingroup
\small
\setlength{\tabcolsep}{9pt}
\renewcommand{\arraystretch}{1.10}
\begin{longtable}{@{}lrrr@{}}
\caption{Initial and final-answer correctness for 371 tasks from one ChatEval GSM8K run. The initial answer is the first Math Expert response. Cells report task counts.}
\label{tab:appendix_chateval_answer_changes} \\
\toprule
\textbf{Initial answer} & \textbf{Final correct} & \textbf{Final incorrect} & \textbf{Total} \\
\midrule
\endfirsthead
\caption[]{\textit{Continued.}} \\
\toprule
\textbf{Initial answer} & \textbf{Final correct} & \textbf{Final incorrect} & \textbf{Total} \\
\midrule
\endhead
\bottomrule
\endlastfoot
Correct & 314 & 36 & 350 \\
Incorrect & 6 & 15 & 21 \\
\midrule
Total & 320 & 51 & 371 \\
\end{longtable}
\endgroup

In Table~\ref{tab:appendix_chateval_answer_changes}, 6 initially incorrect answers become correct, while 36 initially correct answers end with an incorrect final output. We examine one example of each transition below.

\paragraph{A correct answer is lost before final output.}
For the fruit-count task (\texttt{gsm8k-test-917}), the question gives 3 apples, 5 bananas, and 6 oranges, with 2 pieces eaten. The first Math Expert answer is the correct value, 12. Later responses, including the last recorded role answer, also give 12. The final prediction is nevertheless 10, so the correct answer present in the discussion does not reach the final output.

\paragraph{An initial error is corrected.}
For the wage-difference task (\texttt{gsm8k-test-798}), the first Math Expert answer is $-10$. Subsequent responses distinguish one worker's current hourly rate after raises from the other worker's starting rate. The next Math Expert answer is 20; the final prediction retains this value and matches the reference.

\paragraph{Recorded cost on the same subset.}
The 320 tasks with correct final answers use a median of 65,361 recorded tokens, with an interquartile range of 57,851.5--81,678. For the 51 tasks with incorrect final answers, the median is 70,624 tokens and the interquartile range is 58,680--78,303.

%% file: resources/figure/gaia_trace_comparison.tex
\begin{tikzpicture}[
  x=1cm,y=1cm,font=\sffamily\fontsize{10}{12}\selectfont,
  event/.style={draw=black!30,rounded corners=3pt,fill=white,text width=6.65cm,align=left,inner sep=8pt,minimum height=1.4cm},
  flow/.style={-{Stealth[length=2mm]},draw=black!55,line width=.7pt}
]
\definecolor{TraceOrange}{HTML}{AE6024}
\definecolor{TraceBlue}{HTML}{225A97}
\node[align=center,text width=15.2cm,font=\sffamily\bfseries\fontsize{11}{13}\selectfont] at (8,8.25)
  {Task: identify the TV show, restaurant, and book satisfying the two-award-winner condition};
\filldraw[fill=TraceOrange!4,draw=TraceOrange!50,rounded corners=5pt] (0,.75) rectangle (7.75,7.6);
\filldraw[fill=TraceBlue!4,draw=TraceBlue!50,rounded corners=5pt] (8.25,.75) rectangle (16,7.6);
\node[font=\sffamily\bfseries\fontsize{12}{14}\selectfont,text=TraceOrange] at (3.875,7.12) {EvoAgent trace};
\node[font=\sffamily\bfseries\fontsize{12}{14}\selectfont,text=TraceBlue] at (12.125,7.12) {CC-workflow trace};
\node[event] (l1) at (3.875,5.85)
  {{\bfseries Retrieval path}\\[3pt]The search follows the TV show, restaurant, and book clues.};
\node[event] (l2) at (3.875,3.85)
  {{\bfseries Constraint handling}\\[3pt]The answer path drops the two-award-winner condition.};
\node[event] (l3) at (3.875,1.85)
  {{\bfseries Observed outcome}\\[3pt]The final answer does not satisfy the full task condition.};
\node[event] (r1) at (12.125,5.85)
  {{\bfseries Task decomposition}\\[3pt]The subtasks retain the condition attached to the book clue.};
\node[event] (r2) at (12.125,3.85)
  {{\bfseries Evidence handoffs}\\[3pt]Agents exchange candidates and evidence, and consult saved artifacts.};
\node[event] (r3) at (12.125,1.85)
  {{\bfseries Verification and output}\\[3pt]A verifier-stage check occurs before the correct final output.};
\foreach \a/\b in {l1/l2,l2/l3,r1/r2,r2/r3}{\draw[flow] (\a.south) -- (\b.north);}
\node[align=center,text width=15.2cm,font=\sffamily\fontsize{9}{11}\selectfont,text=black!65] at (8,.15)
  {One retained instance: observed events,\\not enforced runtime guarantees.};
\end{tikzpicture}

%% file: resources/paired_gaia_case_study.tex
\subsection{Paired GAIA Response Analysis}
\label{subsec:appendix_paired_gaia}

We retrospectively pair 163 GAIA tasks from separate Jarvis and ChatEval batches. Records must agree in task ID, question, reference, and difficulty, link uniquely to a response within the batch window, and reproduce the result token total from serialized usage. Reference matching is an offline record check, not an inference input. Unmatched records are excluded. This subset is separate from the 165-task main evaluation; the historical batches do not constitute a controlled ablation. The retained ChatEval batch includes responses beyond the two rounds specified in Table~\ref{tab:appendix_reproduced_mas_contracts} and is treated as a separate historical configuration.

\begingroup
\small
\setlength{\tabcolsep}{7pt}
\renewcommand{\arraystretch}{1.10}
\begin{longtable}{@{}lrrr@{}}
\caption{Recorded outcomes and mean tokens for 163 paired GAIA tasks. Correctness labels are retained from the original results.}
\label{tab:appendix_paired_gaia} \\
\toprule
\textbf{Recorded outcome} & \textbf{Tasks} & \textbf{Jarvis tokens} & \textbf{ChatEval tokens} \\
\midrule
\endfirsthead
\caption[]{\textit{Continued.}} \\
\toprule
\textbf{Recorded outcome} & \textbf{Tasks} & \textbf{Jarvis tokens} & \textbf{ChatEval tokens} \\
\midrule
\endhead
\bottomrule
\endlastfoot
Both correct & 44 & 211,388 & 1,538,707 \\
Jarvis only correct & 33 & 253,093 & 1,913,531 \\
ChatEval only correct & 18 & 488,095 & 2,101,207 \\
Both incorrect & 68 & 401,718 & 2,041,633 \\
\end{longtable}
\endgroup

Jarvis and ChatEval have 77 and 62 recorded successes, respectively. ChatEval uses more recorded tokens on 162/163 tasks: its total is $5.72\times$ Jarvis's, and the median within-task ratio is $6.65\times$. Its responses contain 351 explicit code-agent step-limit returns across 127 tasks. These signals describe observed resource use; they do not establish that additional computation causes failure.

\paragraph{Two complementary outcomes.}
We select the task nearest the median ChatEval-minus-Jarvis token difference within each group where only one workflow succeeds, breaking ties by task ID. For the merchant-type task (\nolinkurl{076c8171-9b3b-49b9-a477-244d2a532826}), Jarvis correctly returns \texttt{Finance}. ChatEval's first-round roles and all four Math Expert answers also return \texttt{Finance}, but its recorded final prediction is \texttt{clothing}. The synthesizer output was not retained, so the change cannot be localized to synthesis or answer extraction.

For task \nolinkurl{6b078778-0b90-464d-83f6-59511c811b01}, concerning a bishop\textquotesingle s consecrator, the outcome is reversed: ChatEval's recorded prediction, Alfonso Visconti, is correct, whereas Jarvis returns Giovanni Battista Zeno. ChatEval records 2,409,936 tokens versus Jarvis's 706,050. These cases illustrate both unsuccessful additional processing and a complementary success. They motivate examining stopping rules and the transfer of role answers to final predictions, without establishing the benefits of any intervention.